\documentclass[10pt,twocolumn,letterpaper]{article}
\usepackage[pagenumbers]{iccv}

\usepackage{amsmath,amsfonts,bm}

\def\eqref#1{equation~\ref{#1}}

\def\1{\bm{1}}

\DeclareMathAlphabet{\mathsfit}{\encodingdefault}{\sfdefault}{m}{sl}
\SetMathAlphabet{\mathsfit}{bold}{\encodingdefault}{\sfdefault}{bx}{n}

\usepackage{url}
\usepackage{colortbl} 
\usepackage{booktabs}       
\usepackage{amsfonts} 
\usepackage{enumitem}
\usepackage{nicefrac}       
\usepackage{microtype}      
\usepackage{xcolor}         
\usepackage{graphicx}
\usepackage{amsmath}
\usepackage{multirow}
\usepackage{multicol}
\usepackage{array}
\usepackage{subcaption}
\usepackage{wrapfig}
\usepackage{amsmath,amsfonts,amsthm,amssymb}
\usepackage{bm}

\usepackage[accsupp]{axessibility}  

\usepackage{natbib}

\usepackage{algorithm,algpseudocode}
\algnewcommand{\algorithmicforeach}{\textbf{for each}}
\algdef{SE}[FOR]{ForEach}{EndForEach}[1]
  {\algorithmicforeach\ #1\ \algorithmicdo}
  {\algorithmicend\ \algorithmicforeach}

\usepackage{pifont}

\usepackage{color, colortbl}
\definecolor{Gray}{gray}{0.93}
\definecolor{Orange}{rgb}{1,0.5,0}
\definecolor{DGray}{gray}{0.83}
\definecolor{LightCyan}{rgb}{0.88,1,1}

\usepackage{wrapfig}

\usepackage[dvipsnames]{xcolor}

\usepackage{multirow,mathtools } 
\usepackage{algorithm,algpseudocode}

\usepackage{pifont}
\usepackage{color, colortbl}

\usepackage{blindtext}
\usepackage{lipsum}

\usepackage{multirow}
\usepackage{graphicx}
\usepackage{listings}

\usepackage[most]{tcolorbox}

\definecolor{iccvblue}{rgb}{0.21,0.49,0.74}
\usepackage[pagebackref,breaklinks,colorlinks,allcolors=iccvblue]{hyperref}

\def\paperID{7838} 
\def\confName{ICCV}
\def\confYear{2025}

\title{Towards Stabilized and Efficient Diffusion Transformers through Long-Skip-Connections with Spectral Constraints}

\author{Guanjie Chen\textsuperscript{1,2,6}\footnotemark[1]~\space{}
Xinyu Zhao\textsuperscript{3}\footnotemark[1]~\space{}
Yucheng Zhou\textsuperscript{4}\space{}
Xiaoye Qu \textsuperscript{2,5}\space{}
Tianlong Chen~\textsuperscript{3}\footnotemark[2]\space{}
Yu Cheng\textsuperscript{6}\footnotemark[2]\\\\
\textsuperscript{1}Shanghai Jiao Tong University~~
\textsuperscript{2}Shanghai Artificial Intelligence Laboratory \\
\textsuperscript{3}The University of North Carolina at Chapel Hill~~
\textsuperscript{4}SKL-IOTSC, CIS, University of Macau\\
\textsuperscript{5}Huazhong University of Science and Technology~~
\textsuperscript{6}The Chinese University of Hong Kong \\
\textit{\small chenguanjie@sjtu.edu.cn}, \textit{\small xinyu@cs.unc.edu}
}

\newcommand{\ours}{\texttt{Skip-DiT}}
\newcommand{\ourscache}{\texttt{Skip-Cache}}
\newcommand{\taichi}{\texttt{Taichi}}
\newcommand{\ffs}{\texttt{FFS}}
					
\newcommand{\sky}{\texttt{Sky}}
\begin{document}

\twocolumn[{%
\renewcommand\twocolumn[1][]{#1}%
\maketitle
\begin{center}
    \centering
    \captionsetup{type=figure}
    \vspace{-20pt}
    \includegraphics[width=\linewidth]{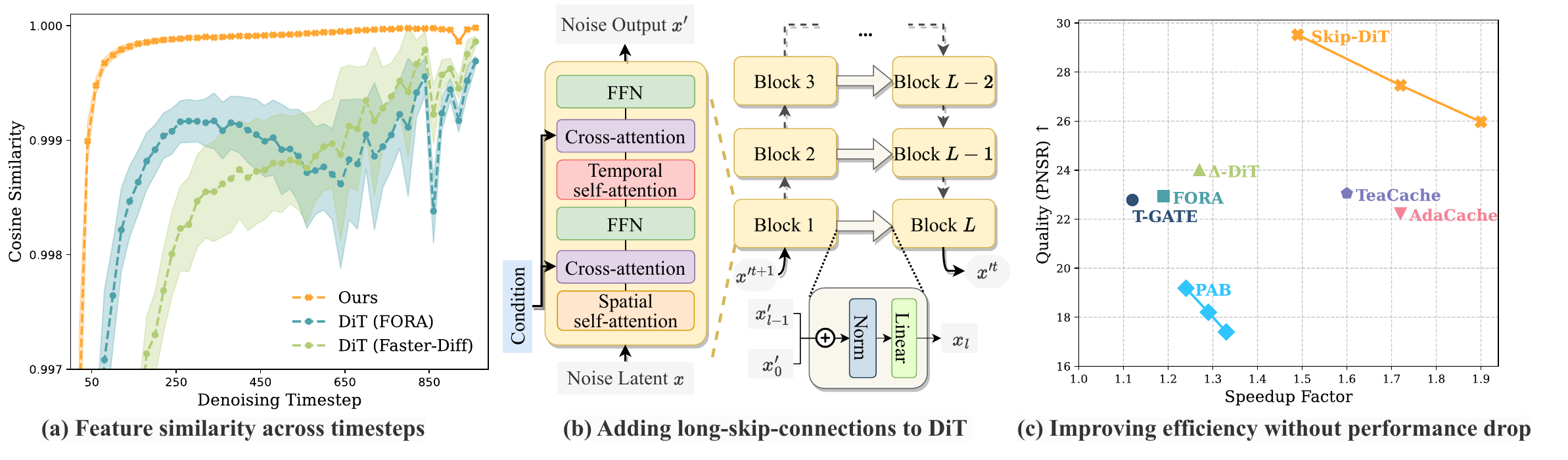}
    \vspace{-18pt}
   \caption{(a) Feature similarities between standard and cache-accelerated outputs in vanilla DiT (caching with FORA~\cite{fora} and Faster-Diff~\cite{li2023fasterdiffusion}) and our proposed \ours. \ours~presents consistently higher feature similarity, demonstrating superior stability after caching. (b) Illustration of \ours~that modifies vanilla DiT models using long-skip-connection to connect shallow and deep DiT blocks. Dashed arrows indicate paths where computation can be skipped in cached inference. (c) Comparison of video generation quality (PNSR) and inference speedup of different DiT caching methods. \ours~maintains higher generation quality even at greater speedup factors.}
    \label{fig:teaser}
    
\end{center}%
}]

\maketitle

\begin{abstract}

Diffusion Transformers (DiT) have emerged as a powerful architecture for image and video generation, offering superior quality and scalability. 
However, their practical application suffers from inherent dynamic feature instability, leading to error amplification during cached inference. 
Through systematic analysis, we identify the absence of long-range feature preservation mechanisms as the root cause of unstable feature propagation and perturbation sensitivity.
To this end, we propose \ours, 
an image and video generative DiT variant enhanced with Long-Skip-Connections (LSCs) - the key efficiency component in U-Nets. Theoretical spectral norm and visualization analysis demonstrate how LSCs stabilize feature dynamics. 
\ours~architecture and its stabilized dynamic feature enable an efficient statical caching mechanism that reuses deep features across timesteps while updating shallow components.
Extensive experiments across the image and video generation tasks demonstrate that \ours~achieves: (1) $4.4\times$ training acceleration and faster convergence, (2) $1.5-2\times$ inference acceleration with negligible quality loss and high fidelity to the original output
, outperforming existing DiT caching methods across various quantitative metrics.
Our findings establish Long-Skip-Connections as critical architectural components for 
stable and efficient diffusion transformers.
Codes are provided in the URL\footnote{\url{https://github.com/OpenSparseLLMs/Skip-DiT}}.
\end{abstract}

\renewcommand{\thefootnote}{\fnsymbol{footnote}}
\footnotetext[1]{Equal Contribution.} 
\footnotetext[2]{Corresponding Authors.}

\section{Introduction}
\label{sec:intro}

Diffusion models~\cite{ddpm,BetkerImprovingIG,Podell2023SDXLIL,zhang2024pia} have emerged as the de-facto solution for visual generation, owing to their high fidelity outputs and ability to incorporate various conditioning signals, particularly natural language. Classical diffusion models typically adopt U-Net~\cite{Ronneberger2015UNetCN} as their denoising backbone. 
Recently, Diffusion Transformers (DiT)~\cite{Chen2023PixArtFT,Peebles2022ScalableDM} introduce an alternative architecture to replace traditional convolutional networks with Vision Transformers, offering enhanced scalability potential. While initially designed for image generation, DiT has demonstrated remarkable effectiveness when extended to video generation tasks~\cite{ma2024latte,li2024hunyuandit,2024arXiv241013720P}. Despite these advances, DiT still faces challenges in practical deployment due to its slow convergence rates~\cite{zhao2024dynamic,yao2025fasterdit,openai2024sora} and substantial inference time requirements, posing significant constraints on training efficiency and real-time applications.


Numerous approaches have been proposed to improve the efficiency of diffusion models, including reduced sampling techniques~\cite{DDIM}, distillation methods~\cite{yin2024one,sauer2023adversarial}, and quantization strategies~\cite{chen2024qdit}. Among these approaches, caching has emerged as one of the most effective strategies for enhancing inference efficiency, owing to its low computational cost while maintaining high fidelity and similarity to the original models through efficient feature reuse~\cite{fora, tgates, delta,adacache,zhao2024pab,teacache}.
However, existing caching methods in DiT still face challenges in visual generation. DiT architecture exhibits significant sample-wise variation in caching function responses, requiring adaptive per-sample hyperparameter tuning and introducing computational overhead~\cite{adacache,teacache}. Besides, current caching methods still yield suboptimal generative results with visible artifacts when cached with larger intervals~\cite{fora, delta, zhao2024pab, adacache}, suggesting error amplification in DiT's transformer blocks. Besides, the training efficiency gap further compounds these challenges. DiT is shown to converge more slowly in training~\cite{zhao2024dynamic,yao2025fasterdit}, and the reasons remain unexplained from architectural perspectives.

DiT efficiency is hindered by the above challenges, but U-Net-based diffusion models can leverage their Long-Skip-Connections (LSCs) to extend cache intervals without complex timestep scheduling or severe performance degradation~\cite{DeepCache,li2023fasterdiffusion}. The LSCs not only make caching stable and robust during inference, but they have also been proven as the key factor for stabilizing training and accelerating convergence for U-Net-based diffusion models~\cite{scalelsc}. These observations lead to our core research question:

\begin{tcolorbox}[before skip=0.2cm, after skip=0.2cm, boxsep=0.0cm, middle=0.1cm, top=0.1cm, bottom=0.1cm]
\textit{\textbf{(Q)} What are the architectural root causes of feature instability and inefficiency in DiT, and to what extent can Long-Skip-Connections mitigate these issues while preserving generation quality?}
\end{tcolorbox}

In this study, we systematically analyze the feature distribution in DiT:
\ding{182} We conduct a preliminary experiment that incorporates Long-Skip-Connections (LSCs) with DiT, and visualize the dynamic features of DiT with and without LSCs by adding perturbations in parameters and model inputs, the feature similarity landscape and plots validate the instability of vanilla DiT (w/o LSCs), which we formally characterize as \textit{Dynamic Feature Instability}, indicating an uncontrolled spectral norm that affects both statistical stability, model robustness, and convergence rate.
\ding{183} We then theoretically prove the superior stability and robustness of DiT with LSCs over vanilla DiT via spectral analysis 

To this end, we propose the Long-\textbf{Skip}-Connected-\textbf{DiT}, namely \ours, 
a DiT architecture that employs spectral constraints—limiting the maximum singular value of weight matrices
to ensure stable gradient flow and reduced sensitivity to perturbations. 
By incorporating these constraints through LSCs, our approach provides theoretical feature stability guarantees and enhanced efficiency in both training and inference, validated through quantitative experiments and visual analysis. 
The LSCs also enable output caching of stabilized features from deep blocks, requiring the computation of only shallow blocks during inference at statically chosen caching timesteps.

To evaluate our proposed \ours, we conduct extensive experiments across 3 DiT backbones, assess training performance on 6 models, and inference efficiency on 7 visual generation tasks, spanning both class and text conditional image/video generation. 
\ours~ achieves superior training efficiency while consistently outperforming both standard baselines and prior caching methods in qualitative and quantitative evaluations.
To summarize, we claim the following contribution of this work:
\begin{itemize}

\item Through theoretical analysis and comprehensive visualization, we identify \textit{Dynamic Feature Instability} as a fundamental root cause that impairs DiT training efficiency and hinders effective inference acceleration.

\item We introduce \ours, an enhanced DiT architecture incorporating LSCs between shallow and deep transformer blocks. We provide theoretical guarantees for feature stability through spectral norm analysis and enable efficient feature caching at strategically selected timesteps.

\item Results of substantial training and inference speedup and nearly undamaged generation quality consistently endorse the effectiveness of \ours. For example, compared to vanilla DiT, \ours~ achieves up to $4.4\times$ training acceleration with even better generation quality. When compared to state-of-the-art DiT caching techniques, \ours~ delivers $1.5$-$2\times$ additional speedup while maintaining superior output fidelity.
\end{itemize}
\section{Related Works}
\label{sec:related_work}
\vspace{-2mm}
\paragraph{Transformer-based Diffusion Models}
The diffusion model has become the dominating architecture for image and video generation, whose main idea is iterative generate high-fidelity images or video frames from noise~\cite{Rombach2021HighResolutionIS}. 
Early diffusion models mainly employ U-Net as their denoising backbone~\cite{Podell2023SDXLIL,BetkerImprovingIG}. However, U-Net architectures struggle to model long-range dependencies due to the local nature of convolutions. Researchers proposing diffusion transformer model (DiT) for image generation~\cite{Chen2023PixArtFT,Bao2022AllAW,Peebles2022ScalableDM}. 
Recent years have witnessed a significant growth in studies of video DiT. Proprietary DiT such as Sora~\cite{openai2024sora} and Movie-Gen~\cite{2024arXiv241013720P} show impressive generation quality, also evidenced by open-sourced implementation~\cite{opensora,pku_opensora_plan}.  Latte decomposes the spatial and temporal dimensions into four efficient variants for handling video tokens, allowing effective modeling of the substantial number of tokens extracted from videos in the latent space~\cite{ma2024latte}.  CogvideoX adds a 3D VAE combined with an expert transformer using adaptive LayerNorm, which enables the generation of longer, high-resolution videos~\cite{yang2024cogvideox}. 
\vspace{-6mm}
\paragraph{Diffusion Acceleration with Feature Caching} 
Since the diffusion model involves iterative denoising, caching features across time-steps, model layers, and modules has been found an effective way to save inference computation costs. For U-Net Diffusion, DeepCache~\cite{DeepCache} and FRDiff~\cite{so2023frdiff} exploit temporal redundancy by reusing features across adjacent denoising steps. While other works take a more structured approach by analyzing and caching specific architectural components--Faster Diffusion~\cite{li2023fasterdiffusion} specifically targets encoder feature reuse while enabling parallel decoder computation, and Block Caching~\cite{wimbauer2024blockcache} introduces automated caching schedules for different network blocks based on their temporal stability patterns. 
Recently, cache-based acceleration has also been applied to DiT. PAB~\cite{zhao2024pab} introduces a pyramid broadcasting strategy for attention outputs. $\Delta$-DiT~\cite{delta} proposes adaptive caching of different DiT blocks based on their roles in a generation--rear blocks during early sampling for details and front blocks during later stages for outlines. T-Gate~\cite{zhang2024cross} identifies a natural two-stage inference process, enabling the caching and reuse of text-oriented semantic features after the initial semantics-planning stage.
To deal with the sample-wise variation in DiT, adaptive caching methods like AdaCache~\cite{adacache} and TeaCache~\cite{teacache} are proposed to predict the feature distribution according to the input. 
\vspace{-3mm}
\paragraph{Long Skip Connections in DiT}
Although previous works like HunyuanDiT \cite{li2024hunyuandit} and U-DiT \cite{u_dit} have incorporated Long Skip Connections (LSCs) into DiTs, a systematic analysis of their functional role has been absent. Our work provides this analysis, highlighting the distinct advantages of our implementation.
We contrast our approach with U-DiT \cite{u_dit}, which employs token and feature downsampling primarily for complexity reduction. Instead, our method uses deep-to-shallow LSCs to enhance the model's spectral properties, boosting feature stability and training convergence. These connections preserve original dimensions, enabling an effective statistic-cached inference by mixing recomputed and cached features. A key advantage is that our method retains the native DiT block structure, permitting conversion from pre-trained models through continued training, unlike U-DiT which necessitates retraining from scratch.

\begin{figure}[thbp]
   \centering
    \vspace{-3mm}
   \begin{subfigure}[b]{1\linewidth}
       \includegraphics[width=\linewidth]{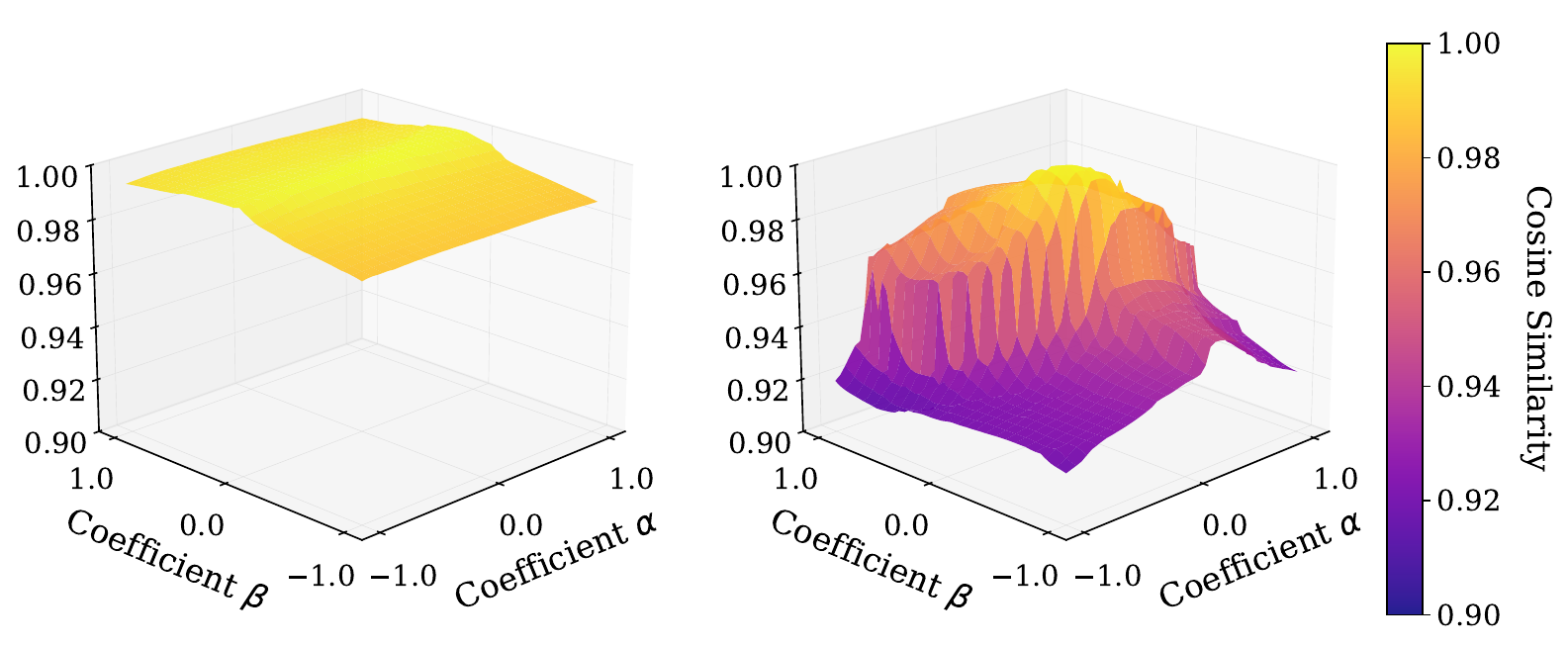}
       \vspace{-3mm}
       \caption{Feature similarity landscape on class-to-image task.}
   \end{subfigure}

   \begin{subfigure}[b]{1\linewidth}
       \includegraphics[width=\linewidth]{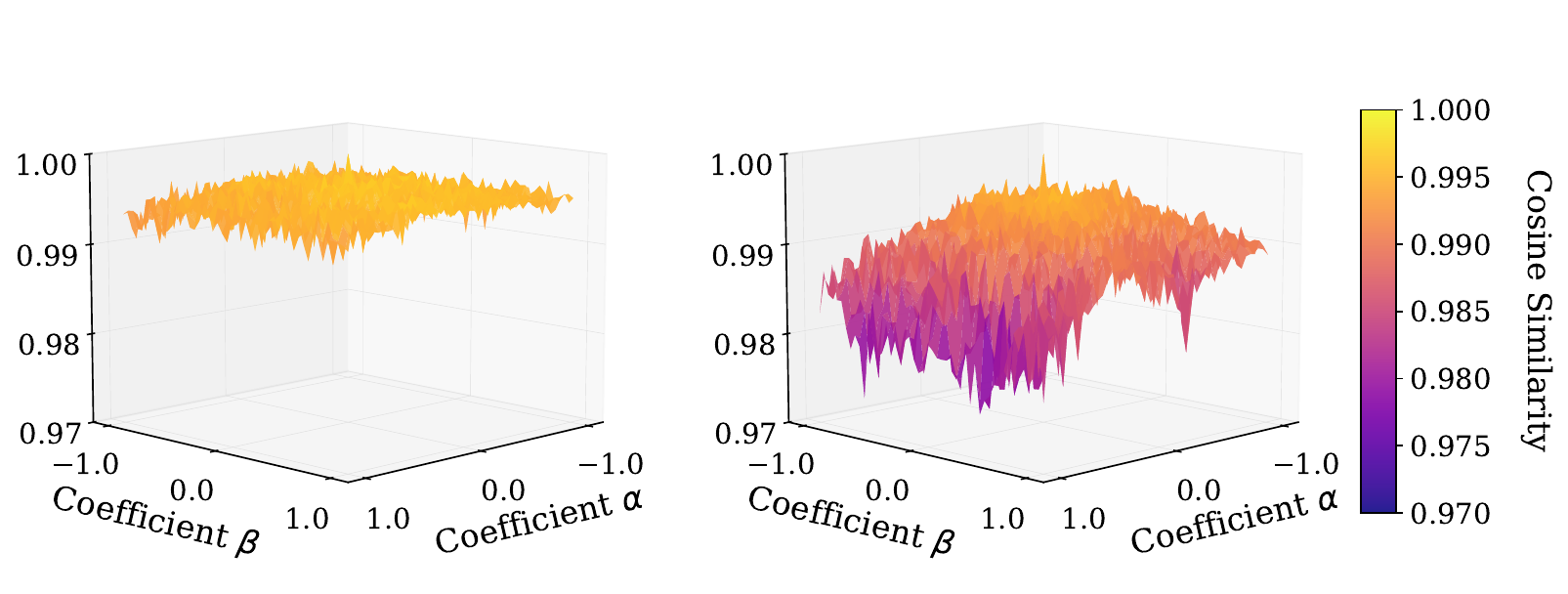}
       \vspace{-3mm}
       \caption{Feature similarity landscape on text-to-video task.}
   \end{subfigure}
   
    \vspace{-2mm}
   \caption{
    Feature stability comparison between \ours~(left) and vanilla DiT (right) on class-to-image (DiT-XL) and text-to-video (Latte) generation. We inject perturbations $\delta$ (magnitudes $\alpha$) and $\eta$ (magnitudes $\beta$), normalized with specific coefficients ($\epsilon = 1e^{-3}$ for DiT-XL and $2e^{-2}$ for Latte), then measure the similarity between the standard and perturbed features.
    }
\label{fig:teaser}
\end{figure}
\vspace{-1.5mm}
\section{Methodology}
\label{sec:methodology}
\vspace{-1.5mm}
\subsection{Preliminaries}
\vspace{-1.5mm}
\paragraph{Diffusion model}
The concept of diffusion models mirrors particle dispersion physics,  where particles spread out with random motion. It involves forward and backward diffusion. The forward phase adds noise to data across $T$ timesteps. Starting from data $\mathbf{x}_0 \sim q(\mathbf{x})$, noise is added to the data at each timestep $t \in \{1\ldots T\}$.
\begin{equation}
\mathbf{x}_t = \sqrt{\alpha_t} \mathbf{x}_{t-1} + \sqrt{1 - \alpha_t} \mathbf{\epsilon}_{t-1}
\end{equation}
where $\alpha$ determines noise level while $\mathbf{\epsilon} \sim \mathcal{N}(0, \mathbf{I})$ represents Gaussian noise. The data $\mathbf{x}_t$ becomes increasingly noisy with time, reaching $\mathcal{N}(0, \mathbf{I})$ at $t = T$. Reverse diffusion then reconstructs the original data as follows, where $\mu_\theta$ and $\Sigma_\theta$ refer to the learnable mean and covariance:
\begin{equation}
p_\theta(\mathbf{x}_{t-1}|\mathbf{x}_t) = \mathcal{N}(\mathbf{x}_{t-1}; \mu_\theta(\mathbf{x}_t, t), \Sigma_\theta(\mathbf{x}_t, t)),
\end{equation}

\subsection{Dynamic Feature Instability in DiT}
\label{sec:instability}
DiT models are known to suffer from low training efficiency~\cite{zhao2024dynamic} and unstable feature distributions during inference~\cite{adacache,teacache}, posing significant challenges for acceleration and their application in interleaved reasoning~\cite{su2025thinking, su2025openthinkimg}. This contrasts sharply with U-Net~\cite{Ronneberger2015UNetCN} architectures where Long-Skip-Connections (LSCs) enhance training stability and efficiency~\cite{scalelsc}, and enable simple but reliable acceleration through feature reuse~\cite{DeepCache,li2023fasterdiffusion}. To explore this architectural discrepancy, we integrate LSCs into DiT-XL~\cite{dit-xl} as in Figure~\ref{fig:teaser} (b) and train from scratch as the preliminary experiment. Training settings is the same as in Section \ref{c2i-train-detail} 

\begin{figure}[!t]
   \centering
   \vspace{-4mm}
   \begin{subfigure}[b]{0.48\linewidth}
       \includegraphics[width=\linewidth]{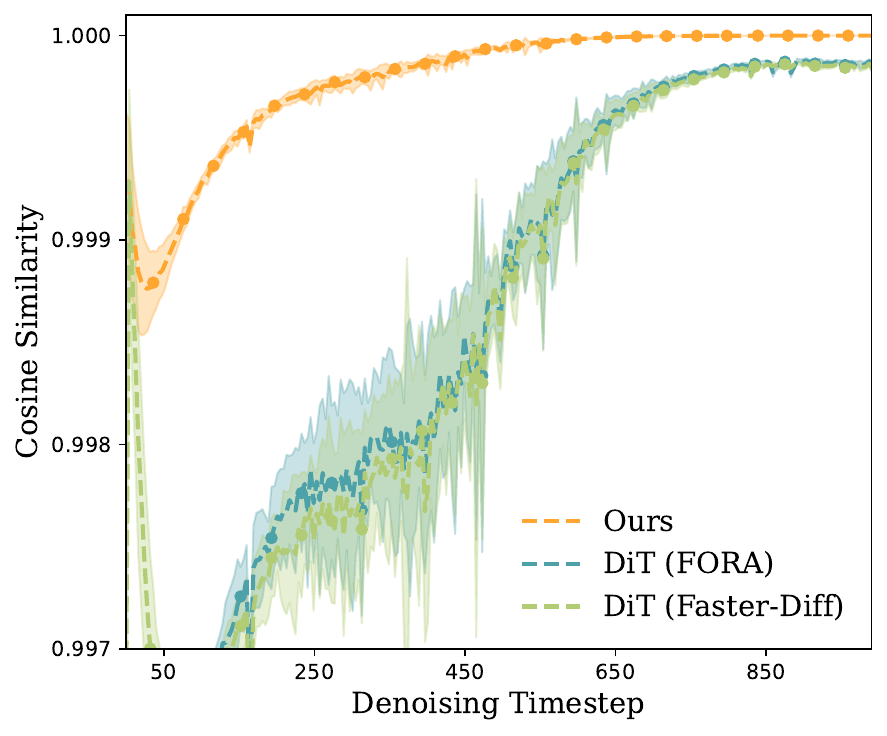}
       \caption{Class-to-image Task.}
   \end{subfigure}
   \hfill
   \begin{subfigure}[b]{0.48\linewidth}
       \includegraphics[width=\linewidth]{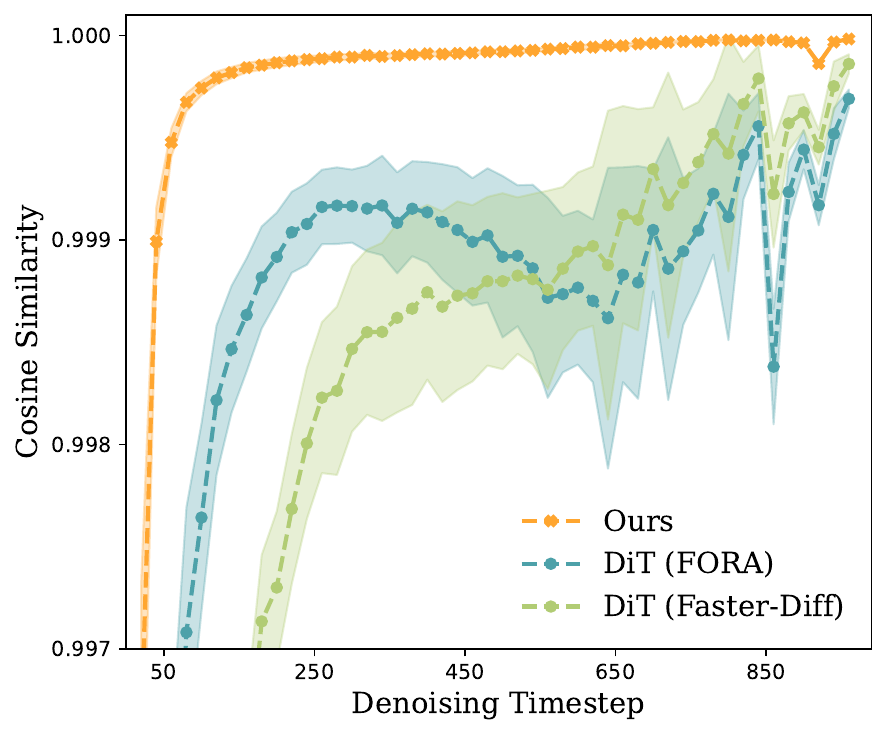}
       \caption{Text-to-video Task.}
   \end{subfigure}
    \vspace{-2mm}
   \caption{
    Comparison between standard and cache-accelerated outputs in vanilla DiT versus \ours, with Latte (text-to-video generation) and DiT-XL (class-to-image generation) serving as base architectures. Both mean and standard deviation across samples are shown. Vanilla DiT exhibits much higher sample variance.
    }
    \vspace{-3mm}
\label{fig:teaser2}
\end{figure}

\vspace{-2mm}
\paragraph{Visualizing Feature Stability of DiT}
Let $\theta \in \mathbb{R}^d$ denote the original parameters of model $f(x;\theta)$. Given random perturbation vectors $\delta, \eta \in \mathbb{R}^d$, $\|\delta\|=\|\eta\|=\epsilon\|\theta\|$, with $0<\epsilon<1$ which is normalization coefficient. we construct perturbed parameters:
$\theta' = \theta + \alpha\delta + \beta\eta$,
where $\alpha, \beta \in \mathbb{R}$ control perturbation magnitudes. The feature stability landscape for 3D visualization is defined as:
\begin{equation}
    L(\alpha,\beta) = \mathrm{CosineSimilarity}\left(f(x;\theta), f(x;\theta')\right)
    \label{eq:landscape}
\end{equation}
To visualize the feature stability across timesteps, we perturb the inputs with cached values, following the caching strategy of FORA\cite{fora} and FasterDiff\cite{yao2025fasterdit}, and we take cosine similarity to measure the distance with and without caching. Visualized results are illustrated in Figure~\ref{fig:teaser}(a) and \ref{fig:teaser2}(a).

\vspace{-2mm}
\paragraph{Dynamic Feature Instability}
Our preliminary experiments and visualizations reveal that vanilla DiT models exhibit significant sensitivity to perturbations in both parameters and input features, with feature similarity across timesteps demonstrating considerable variability across different examples. These observations raise critical concerns regarding the stability of DiT models, particularly for training efficiency and inference acceleration. We formally define the instability of DiT as  \textit{Dynamic Feature Instability} (DFI).
We preliminary attribute these issues to the uncontrolled spectral norm $\sigma_{\max}$ of DiT’s Jacobians $J^{\text{vanilla}}$, which exponentially amplifies perturbations 
and destabilizes feature propagation. This aligns with findings in~\cite{SpectralNormRegularization}, where high spectral norms degrade model robustness and generalization capacity by amplifying input sensitivities, and mirrors the instability patterns observed in visual generative GAN training~\cite{SpectralNormalizationGAN} without spectral normalization.

\begin{figure}[t]
    \centering
    \vspace{-4mm}
    \includegraphics[width=\linewidth]{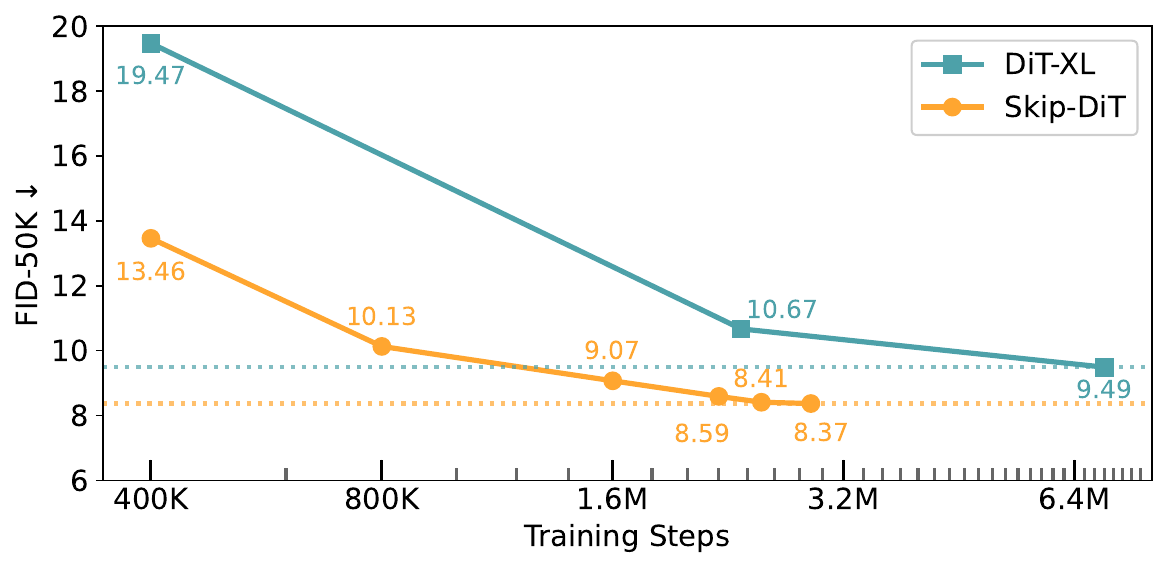}
    \vspace{-6mm}
    \caption{
    Training efficiency comparison between \ours~ and DiT-XL. \ours~achieves superior FID-50K score on ImageNet at 1.6M training steps (vs. DiT-XL’s 7M steps) and converges faster at 2.9M steps. Both models are evaluated on FID-50K under identical settings (no classifier-free guidance, cfg=1).
    }
    \vspace{-4mm}
    \label{fig:train-compare}
\end{figure}

\vspace{-2mm}
\paragraph{Training Convergence Comparison}
To validate the theoretical convergence rate in real implementations, we extend preliminary experiments recording the relationship between training steps and performance and FID\cite{fid} on ImageNet\cite{deng2009imagenet}. Both DiT models with/without LSCs are trained from scratch under identical settings. Results are revealed in Figure \ref{fig:train-compare}. 
The accelerated convergence of DiT w/ LSCs observed in practical training demonstrates the training efficiency and model capability. This mirrors our theoretical analysis in Theorem~\ref{theorem1}, confirming that the spectral norm is effectively controlled with LSCs during training.

\subsection{Integerating Long-Skip-Connection into DiT}
\label{sec:skip-dit}
To detail the DiT with Long-Skip-Connections(LSCs), we connect the shallow blocks to deep blocks with LSCs in DiT, named as \ours.
Let $\boldsymbol{x}$ denote the input noise embedding, and $\boldsymbol{x}'_l$ represents the output at the $l$-th layer of \ours. The architecture consists of $L$ sequential DiT blocks with LSCs. 
Each DiT block $\mathcal{F}_\text{DiT}^l$ at block $l$ processes the features as $\boldsymbol{x}' = \mathcal{F}_\text{DiT}^l(\boldsymbol{x})$.
The $i$-th skip branch ($i \in \{1\ldots L//2\}$) connects $i$-th block to ($L+1-i$)-th block, which can be denoted as $\mathcal{F}_\text{Skip}^i(\cdot,\cdot)$. Given output $\boldsymbol{x}'_{i}$ from the start of the skip branch and $\boldsymbol{x}'_l$ from the previous layer, the skip branch aggregates them to the input to $l$-th block as:
\begin{align}
\boldsymbol{x}_l = \mathcal{F}_\text{skip}^i(\boldsymbol{x}'_{i} , \boldsymbol{x}'_{l-1}) = \texttt{Linear}(\texttt{Norm}(\boldsymbol{x}'_{i} \oplus \boldsymbol{x}'_{l-1}))
\end{align}
where $\oplus$ denotes concatenation, $\texttt{Norm}$ represents the layer normalization, and $\texttt{Linear}$ is a fully-connected layer. Each skip connection creates a shortcut path that helps preserve and process information from earlier layers, enabling better gradient flow and feature reuse throughout the network. 

\subsection{Spectral Analysis of \ours}
\label{idea_model}
\newtheorem{theorem}{Theorem}

Consider an ideal denoising diffusion transformer $M$ with $L$ identical blocks, where the denoising capability induces the following fundamental properties:

\begin{itemize}[leftmargin=*]
\setlength{\itemsep}{0pt} 
\item \textit{Noise Reduction Invariance}: Each transformer block $\mathcal{T}_l$ strictly reduces noise magnitude of input $h$. This directly implies the Jacobian spectral norm ($\sigma_{max}$) constraint:
    \begin{equation}
    \sigma_{\max}\left( \frac{\partial \mathcal{T}_l}{\partial h} \right) \triangleq \gamma_l < 1
    \end{equation}
    
    \item \textit{Transformer Blocks Homogeneity}: Identical noise reduction ratio across layers
    $\gamma_l = \gamma,\quad \forall l \in \{1,...,L\}$. Thus, the complete model satisfies
\vspace{-1mm}
\begin{equation}
    \sigma_{\max}(M) = \prod_{l=1}^L \gamma = \gamma^L \ll 1
\end{equation}
\end{itemize}


\begin{theorem}
The spectral norm of the Jacobian matrix of DiT with Long-Skip-Connections is controlled more tightly than that of Vanilla DiT $M$, making the Skip-DiT model more robust, numerically stable, and capable of converging faster.

\label{theorem1}
\end{theorem}

\noindent
\textit{Proof.}~
Define layer transformations for \( L/2 < l \leq L \) and their Jacobian matrices, 
where \( \mathcal{T} \) denotes the Transformer block and \( 0 < \alpha < 1 \) denotes normlized concatetion:
\begin{align*}
h_{l+1}^{\text{vanilla}} = \mathcal{T}(h_l^{\text{vanilla}}), h_{l+1}^{\text{skip}} = (1-\alpha) \cdot \mathcal{T}(h_l^{\text{skip}}) + \alpha \cdot h_{L-l}^{\text{skip}}.
\end{align*}
The corresponding Jacobian matrices $J$ are:
\begin{align*}
J_{l}^{\text{vanilla}} = \frac{\partial \mathcal{T}\left(h_l\right)}{\partial h_l}, J_{l}^{\text{skip}} = (1-\alpha) \cdot \frac{\partial \mathcal{T}\left(h_l\right)}{\partial h_l} + \alpha \cdot \frac{\partial h_{L-l}}{\partial h_l}.
\end{align*}
Given $\gamma < 1$ and $2l-L \geq 1$ for $l \geq L/2$, the layer wise spectral norms $\sigma_{max}$ of $M_{vanilla}$ and $M_{skip}$ staisfy:
\begin{align*}
\sigma_{\max}(J_{l}^{\mathrm{skip}}) &\leq (1-\alpha)\gamma + \alpha\gamma^{2l-L}  < \gamma = \sigma_{\max}(J_{l}^{\mathrm{vanilla}})
\end{align*}
For the full model Jacobian $J = \prod_{l=1}^{L} J_l$, the $\sigma_{max}$ satisfy:
\begin{align}
\sigma_{\max}(J^{\mathrm{skip}}) &< \prod_{l=0}^{L-1} \gamma = \gamma^L = \sigma_{\max}(J^{\mathrm{vanilla}}).
\end{align}
\noindent
This spectral norm reduction attenuates perturbation growth through $\| \Delta h_{L} \| \leq (\sigma_{\max}(J))^L \| \Delta h_0 \|$, stabilizes gradient flow via $\| \nabla_{\theta}\mathcal{L} \| \leq \sigma_{\max}(J) \| \nabla_{h_L}\mathcal{L} \|$.
This theoretical analysis aligns with our findings in Section~\ref{sec:instability}, where \ours~ is proven to be more feature stable and training efficient through visualizations and preliminary experiments.

\subsection{Cached Inference Mechanism}
The feature stability and robustness in \ours~enables effective reuse of intermediate features across timesteps during denoising. To quantify this advantage, we analyze the similarity of transformer block outputs with versus without caching in both vanilla DiT and \ours.
\label{skip-cache}
\vspace{-3mm}
\paragraph{Caching Strategy for \ours}
For \ours, we implement a hybrid caching strategy combining skip-connection-based block skipping from DeepCache~\cite{DeepCache}, phased denoising strategy\cite{li2023fasterdiffusion,zhang2024cross}, and errors accumulation guidance\cite{adacache,teacache}. As illustrated in Figure \ref{fig:teaser2}, our guidance mechanism is informed by a static error record $\mathcal{R}$, which is pre-computed by averaging historical prediction errors. 
To optimize the caching process, we also identify an optimal dynamic phase, $\mathcal{P}$, through static feature analysis. The choice $\mathcal{P}$ is substantiated by our ablation study in Table \ref{tab:ablation_timestep} and visualization in Figure \ref{fig:heatmap2}. 
The complete inference procedure is formally outlined in Algorithm \ref{alg:cached_inference}. 
\vspace{-2mm}
\paragraph{Visualizing Caching Stability} As visualized in Figure~\ref{fig:teaser2}, \ours~ exhibits significantly higher feature similarity and smaller variance among examples compared to vanilla DiT, which employs FORA~\cite{fora} and FasterDiffusion designed for DiT models. This demonstrates superior stability for cached inference. We attribute this to the inherent feature stabilization and enhanced robustness. The latter can be theoretically grounded on $M$ defined in Section \ref{idea_model}:
\begin{theorem}
Under feature reuse with interval $\tau$, Skip-DiT achieves tighter perturbation bounds than Vanilla DiT, enabling larger allowable reuse intervals while maintaining error tolerance $\epsilon_{\mathrm{max}}$.
\end{theorem}

\noindent
\textit{Proof.}~
Let $h_{t}$ be the reused feature with error $\epsilon_t$, $\delta$ the perturbation bound per step, $x$ the input, and $\theta$ the model parameters. Through the differential mean-value theorem and error propagation:
\begin{align*}
\epsilon_{t+1}=\| h_{t+1}-h_t \| &\leq \mathrm{Lip}(M)\epsilon_t + \delta
\end{align*} 
From Theorem~\ref{theorem1}, the Lipschitz constants and perturbation bounds satisfy:
$\mathrm{Lip}(M_{\mathrm{skip}}) < \mathrm{Lip}(M_{\mathrm{vanilla}})$ and $\delta_{\mathrm{skip}} < \delta_{\mathrm{vanilla}}$.
For $T$-step feature reuse, the cumulative error is:
\begin{equation}
\epsilon_T \leq \frac{\mathrm{Lip}(M)^T - 1}{\mathrm{Lip}(M) - 1}\delta \label{eq:cumulative_error}
\end{equation}
Considering the conditions above, the maximal interval $\tau$ satisfies: $\epsilon_\tau^{skip}=\epsilon_\tau^{vanilla} < \epsilon_{max}$.
Solving this yields $\tau_{\mathrm{skip}} > \tau_{\mathrm{vanilla}}$ under identical $\epsilon_{\mathrm{max}}$, which demonstrate better robustness and tolerance to error of \ours.

\renewcommand{\algorithmicrequire}{\textbf{Input}}
\renewcommand{\algorithmicensure}{\textbf{Output}}

\begin{algorithm}[t]
\caption{\ours~ with Cached Inference}
\label{alg:cached_inference}
\begin{algorithmic}[1]
\Require Initial state $\boldsymbol{x}^t$, model $\mathcal{F}$, cache interval $N$, global error $\epsilon$, static error records $\mathcal{R}(x_{t=1}^{T})$, dynamic Phase $\mathcal{P}$.
\Ensure Generated sequence $\{\boldsymbol{x}^{t}\}_{t=1}^T$

\State \textbf{Global Timestep $t$ (Full Inference)}:
\State $\boldsymbol{x'}^t \leftarrow \mathcal{F}^{L-1} \circ \cdots \circ \mathcal{F}^1(\boldsymbol{x}^t)$
\State Cache feature: $\mathcal{C}^t_{L-1} \leftarrow \boldsymbol{x'}^t$
\State $\boldsymbol{x}^{t-1} = \mathcal{F}^L(\boldsymbol{x}^t)$
\For{$k = 1$ \textbf{to} $N-1$}
    \State \textbf{Local Timestep $t-k$ (Cached Inference)}:
    \State $\boldsymbol{x}'^{t-k}_1 \leftarrow \mathcal{F}_\text{DiT}^1(\boldsymbol{x}^{t-k})$
    \State $\boldsymbol{x}^{t-k} \leftarrow \mathcal{F}_\text{DiT}^L\left(\mathcal{F}_\text{Skip}^1(\boldsymbol{x}'^{t-k}_1, \mathcal{C}^t_{L-1})\right)$
    \State Maintain cache: $\mathcal{C}^t_{L-1} \leftarrow \mathcal{C}^t_{L-1}$
    \State Accumulate error: $\epsilon \leftarrow \epsilon + \mathcal{R}(t-k)$
    \If{$t-k-1 \in \mathcal{P} ~||~\epsilon>\text{threshold}$} 
    \State \textbf{$\epsilon\leftarrow0$; Break}
\EndIf
\EndFor
\State \textbf{$t\leftarrow t-k-1$}; 
\State \textbf{Back to Full Inference until: $t<0$;}
\end{algorithmic}
\end{algorithm}


\section{Experiments}
\label{sec:experiments}
\begin{table*}[th]
    \caption{
    Class-to-video generation performance. The definition of the cache step $n$ follows that in Table \ref{tab:t2v}. For the UCF101 task, the scheduler is 50-steps-DDIM, while for the remaining tasks is 250-steps-DDPM. Latency and speedup are calculated on one A100 GPU. The notation $n=i$ represents cache interval $N$ in Algorithm \ref{alg:cached_inference}. We highlight Baseline DiT models in \colorbox{gray!20}{grey}, and the best metrics in \textbf{bold}.
    }
    \label{tab:c2v}
    \scriptsize \centering
    \setlength{\tabcolsep}{7.2pt}
    \begin{tabular}{clcccccccccccc}
    \toprule
    \multirow{2}{*}{Method} & \multicolumn{2}{c}{\texttt{UCF101}} & \multicolumn{2}{c}{\ffs} & \multicolumn{2}{c}{\sky} & \multicolumn{2}{c}{\taichi} & \multirow{2}{*}{FLOPs (T)} & \multirow{2}{*}{Latency (s)} & \multirow{2}{*}{Speedup} \\
    & FVD ($\downarrow$) & FID ($\downarrow$) & FVD ($\downarrow$) & FID ($\downarrow$) & FVD ($\downarrow$) & FID ($\downarrow$) & FVD ($\downarrow$) & FID ($\downarrow$) & & & \\
    \midrule
    \rowcolor[gray]{0.9} 
    Latte & 155.22 &	22.97 &	28.88 &	5.36 &	49.46 &	11.51 &	166.84 & 11.57 & 
    278.63 & 9.90 & 1.00$\times$ \\
        \cmidrule{1-12}
        $\Delta$-DiT & 161.62 &	25.33 &	25.80 &	4.46 &	51.70 &	\textbf{11.67} &	188.39 & \textbf{12.09} & 
     226.10 & 8.09 & 1.22$\times$ \\
        FORA & 160.52 &	23.52 &	27.23	& 4.64	& 52.90	& 11.96	& 198.56	& 13.68 &
    240.26 & 9.00 & 1.10$\times$ \\
    \cmidrule{1-12}
        PAB$_{23}$ & 213.50 &	30.96 &	58.15 &	5.94 &	96.97 &	16.38	& 274.90 &	16.05 &
    233.87 & 7.63 & 1.30$\times$ \\
        PAB$_{35}$ & 1176.57 &	93.30 &	863.18 &	128.34 &	573.72	& 55.66	& 828.40	& 42.96 & 
    222.90 & 7.14 & 1.39$\times$ \\
        \cmidrule{1-12}			
        
    \multicolumn{12}{c}{\ours~ with Cached Inference} \\ \cmidrule{1-12}
    \rowcolor[gray]{0.9} 
    original & 141.30 &	23.78 &	20.62 &	4.32 &	49.21 &	11.92 &	163.03 & 13.55 & 
    290.05 & 10.02 & 1.00$\times$ \\
    \cmidrule{1-12}
        $n=2$ & 141.42 &	21.46 &	\textbf{23.55} &	\textbf{4.49} &	\textbf{51.13} &	12.66 &	\textbf{167.54} &	13.89 (\textbf{0.34$\uparrow$}) & 180.68 & 6.40 & 1.56$\times$ \\
        $n=3$ & \textbf{137.98}	& 19.93 &	26.76 &	4.75 &	54.17 &	13.11 &	179.43 & 14.53 & 145.87 & 5.24 & 1.91$\times$ \\
        $n=4$ & 143.00	& 19.03 &	30.19 &	5.18 &	57.36 &	13.77 &	188.44 &	14.38 & 125.99 & 4.57 & 2.19$\times$ \\
        $n=5$ & 145.39 &	\textbf{18.72} &	35.52 &	5.86 &	62.92 &	14.18 &	209.38 &	15.20 & 121.02 & 4.35  & 2.30$\times$ \\
        $n=6$ & 151.77 &	18.78 &	42.41 &	6.42 &	68.96 &	15.16 &	208.04 &	15.78 & \textbf{111.07} & \textbf{4.12} & \textbf{2.43}$\times$ \\
        \bottomrule
    \end{tabular}
\end{table*}

\subsection{Implementation Details}
\paragraph{Models}
To start with, we add LSCs in DiT-XL~\cite{dit-xl} on the class-to-image task to prove the training efficiency and feature stability of ~\ours. Then we employ Latte~\cite{ma2024latte} as our base model in video generation and Hunyuan-DiT for text-to-image generation to further demonstrate the remarkable effectiveness of \ours~ and its caching efficiency. All DiT models are integrated with LSCs to evaluate the performance in both video and image generation tasks following the guidance in Section.\ref{sec:skip-dit} and explore techniques for integrating LSCs into a pre-trained DiT model.
\vspace{-2mm}
\paragraph{Datasets}
For the class-to-image task, \ours~is trained on 256$\times$256 ImageNet~\cite{deng2009imagenet}.
In the class-to-video task, \ours~ is trained on four datasets: FaceForensics~\cite{ffs}, SkyTimelapse~\cite{sky}, UCF101~\cite{ucf101}, and Taichi-HD~\cite{taichi}. Following the experimental settings in Latte, we extract 16-frame video clips from these datasets and resize all frames to a resolution of 256$\times$256. 
For the text-to-video task, the original Latte is trained on Webvid~\cite{webvid} and Vimeo~\cite{vimeo}, comprising approximately 330k text-video pairs in total. Considering that the resolution of Webvid is lower than 512$\times$512, we only sample 330k text-video pairs from Vimeo for training, with a resolution of 512$\times$512 and 16 frames.
\vspace{-2mm}
\paragraph{Training Details}
\label{c2i-train-detail}
In the class-to-image task, the model is initialized randomly and aligns all the training settings with DiT-XL. The detailed training settings can be found in the Appendix.
For class-to-video tasks, we train all instances of \ours~ from scratch on all datasets.
For text-to-video tasks, we propose a two-stage continual training strategy:
\begin{itemize}[leftmargin=*]
\setlength{\itemsep}{0pt} \item \textit{Skip-Connections training:} Latte with LSC is initialized with the weights of the original Latte, and the weight of LSCs is initialized randomly. At this stage, only the LSCs are trained. It only takes one day on 8 H100 GPUs.
    \item \textit{Overall training:} With LSCs Fully trained, all other parameters are unfrozen to perform overall training. At this stage, \ours~rapidly recovers its generation capability within two days and can generate content comparable to the original Latte with an extra three days of training. 
\end{itemize}
All our training experiments are conducted on 8 H100 GPUs, video-image joint training strategy proposed by~\citet{ma2024latte} is employed to enhance training stability.


\subsection{Results on Video Generation Tasks}
\paragraph{Evaluation Details}
Following previous works~\cite{zhao2024pab, adacache, teacache} and Latte, we evaluate text-to-video models using \texttt{VBench}~\cite{VBench}, Peak Signal-to-Noise Ratio (PSNR), Learned Perceptual Image Patch Similarity (LPIPS)~\cite{lpips}, and Structural Similarity Index Measure (SSIM)~\cite{ssim}. \texttt{VBench} is a comprehensive benchmark suite comprising 16 evaluation dimensions. PSNR is a widely used metric for assessing the quality of image reconstruction, LPIPS measures feature distances extracted from pre-trained networks, and SSIM evaluates structural information differences. 
For class-to-image tasks, we evaluate the similarity between generated and real videos using Fréchet Video Distance (FVD)~\cite{fvd} and Fréchet Inception Distance (FID)~\cite{fid}, following the evaluation guidelines of StyleGAN-V~\cite{stylegenv}. 

\newcommand{\graycell}[1]{\cellcolor[gray]{0.9} #1}

\begin{table*}[t]
    \scriptsize \centering
    \caption{
    Text-to-video generation performance. Inference latency and flops are evaluated with 1 A100 GPU. The notation $n=i$ represents the cache interval $N$ in Algorithm \ref{alg:cached_inference}.
    Baseline DiT models are highlighted in \colorbox{gray!20}{grey}, and the best metrics are emphasized in \textbf{bold}.
}
    \label{tab:t2v}
    \setlength{\tabcolsep}{9pt}
    \begin{tabular}{clccccccc}
        \toprule
        Model & Method & VBench(\%) $\uparrow$ & PSNR $\uparrow$  & LPIPS $\downarrow$ & SSIM $\uparrow$ & FLOPs (T) & Latency (s) & Speedup \\
        \midrule

         \multirow{9}{*}{\textbf{Latte}} & \graycell{original} & \graycell{76.14} &	\graycell{--} &	\graycell{--} & \graycell{--} & 
    \graycell{1587.25} & \graycell{27.11} & \graycell{1.00$\times$}  \\
        \cmidrule{2-9}
        & T-GATE & 75.68 ~~($\downarrow$0.46) &	22.78 &	0.19 & 0.78 & 
    1470.72 & 24.15 & 1.12$\times$ \\
        & FORA & 76.06 ~~($\downarrow$0.08) &	22.93 &	0.14 & 0.79 & 
    1341.72 & 24.21 & 1.19$\times$ \\
        & $\Delta$-DiT & 76.06 ~~($\downarrow$0.08) &	24.01 &	0.17 & 0.81 & 
    1274.36 & 21.40 & 1.27$\times$ \\
        & PAB$_{235}$ & 73.77 ~~($\downarrow$2.37) &	19.18 &	0.27 &	0.66 & 1288.08 & 23.24 & 1.24$\times$ \\
        & PAB$_{347}$ & 71.11 ~~($\downarrow$5.03) &	18.20 &	0.32 &	0.63 & 1239.35 & 22.23 & 1.29$\times$ \\
        & PAB$_{469}$ & 70.20 ~~($\downarrow$5.94) &	17.40 &	0.35 &	0.60 & 1210.11 & 21.60 & 1.33$\times$ \\
        & AdaCache & \textbf{76.07} ~~($\downarrow$0.07) &	22.23 &	0.19 &	0.78 & 934.85 & 16.95 & 1.60$\times$ \\
        & TeaCache & 76.03 ~~($\downarrow$0.11) &	23.05 & 0.17 &	0.80 & 862.52 & 15.75 & 1.72$\times$ \\
        \cmidrule{1-9}
        
        \multirow{11}{*}{\textbf{\ours}} & \graycell{original} & \graycell{75.60} &	\graycell{--} &	\graycell{--} & \graycell{--} & 
        \graycell{1648.13} & \graycell{28.72} & \graycell{1.00$\times$}  \\
        
        \cmidrule{2-9}
        & \multicolumn{8}{c}{$65\%$ steps cached} \\  \cmidrule{2-9}
        & $n=2$ & 75.51~~($\downarrow$0.09) &	\textbf{29.52} &	\textbf{0.06} &	\textbf{0.89} & 1127.83 & 19.28 & 1.49$\times$ \\
        & $n=3$ & 75.26~~($\downarrow$0.34) &	27.46 &	0.09 &	0.85 & 974.80 & 16.67 & 1.72$\times$ \\
        & $n=4$ & 74.73~~($\downarrow$0.87) &	25.97 &	0.13 &	0.81 & 882.98 & 15.12 & 1.90$\times$ \\

        \cmidrule{2-9}
        & \multicolumn{8}{c}{$75\%$ steps cached} \\  \cmidrule{2-9}
        & $n=2$ & 75.36~~($\downarrow$0.24) &	26.02 &	0.10 &	0.84 & 1066.62 & 18.25 & 1.57$\times$ \\
        & $n=3$ & 75.07~~($\downarrow$0.53) &	22.85 &	0.18 &	0.76 & 852.38 & 14.88 & 1.93$\times$ \\
        & $n=4$ & 74.43~~($\downarrow$1.17) &	22.08 &	0.22 &	0.73 & \textbf{760.56} & \textbf{13.03} & \textbf{2.20}$\times$ \\

        \bottomrule
    \end{tabular}
    
\end{table*}

\vspace{-3mm}
\paragraph{Class-to-video Generation}
We compare the quantitative performance of Latte and \ours~on four class-to-video tasks, as shown in Table \ref{tab:c2v}. \ours~consistently outperforms Latte in terms of FVD scores across all tasks while achieving comparable performance in FID scores, demonstrating its strong video generation capabilities. Furthermore, we observe that \ours~ with cached inference significantly outperforms other caching methods across most metrics, incurring only an average loss of $2.37$ in the FVD score and $0.27$ in the FID score while achieving a 1.56$\times$ speedup. 
\paragraph{Text-to-video Generation}
\vspace{-2mm}
Table \ref{tab:t2v} presents evaluation results on text-to-video models. Videos are generated using the prompts of VBench~\cite{VBench}, which is considered a more generalized benchmark~\cite{zhao2024pab,VBench,opensora}. Compared with the original Latte, \ours~achieves a comparable \texttt{VBench} score (75.60 \textit{vs.} 76.14) with only six days of continual pre-training on 330k training samples. To demonstrate the superiority of the cached inference of \ours~, we evaluate two caching settings: caching at timesteps $700\rightarrow50$ and $800\rightarrow50$ (out of 1000 timesteps in total). In both settings, \ours~achieves the highest speedup while maintaining superior scores in PSNR, LPIPS, and SSIM, with only a minor loss in \texttt{VBench} score. 
\subsection{Results on Image Generation Tasks}
\begin{table*}[ht]
    \caption{
        Text-to-image generation performance. Latency and speedup are calculated on one A100 GPU. The notation $n=i$ represents cache interval $N$ in Algorithm \ref{alg:cached_inference}. We highlight baseline DiT models (without caching) in \colorbox{gray!20}{grey}, and the best metrics in \textbf{bold}.
    }
    \label{tab:t2i}
    \scriptsize \centering
    \setlength{\tabcolsep}{10pt}
    \begin{tabular}{clccccccc}
    \toprule
    Method
    & FID ($\downarrow$) & CLIP ($\uparrow$) & PSNR ($\uparrow$) & LPIPS ($\downarrow$) & SSIM ($\uparrow$) & FLOPs (T) & Latency (s) & Speedup \\
    \midrule
    \rowcolor[gray]{0.9} 
    HunYuan-DiT & 32.64 &	30.51 &	-- &	-- &	-- &	514.02 &	18.69 & 1.00 \\
        \cmidrule{1-9}
        
        TGATE & 32.71 & 30.64 & 16.80 &	0.24 &	0.61 &	378.94 &	13.21 & 1.41 \\
        
        $\Delta$-Cache & 28.35 &	30.35	& 16.56 &	0.21	& 0.65	& 362.67	& 13.58 & 1.38 \\
        
        FORA & 31.21	& 30.53	& 19.58	& 0.14	& 0.75	& 330.68	& 13.20 & 1.42 \\
        \cmidrule{1-9}
        \multicolumn{9}{c}{\ours~ with Cached Inference} \\  \cmidrule{1-9}
        $n=2$ & 31.30 &	30.52	& \textbf{22.09} &	\textbf{0.10} &	\textbf{0.84}	&348.24 &	12.76 &	1.46 \\

        $n=3$ & 29.53 &	30.55 &	21.25 &	0.11 &	0.81 &	299.48 &	10.91 &	1.71 \\

        $n=4$ & 27.49 &	30.55 &	20.55 &	0.13 &	0.78 &	270.22 &	10.02	& 1.87 \\

        $n=5$ & 28.37 &	30.56 &	19.94 &	0.14 &	0.76 &	260.47	& 9.51 &	1.96 \\
        $n=6$ & \textbf{27.21} &	\textbf{30.71} &	19.18 &	0.18 &	0.70 &	\textbf{240.96}	& \textbf{8.96} &	\textbf{2.09} \\




        \bottomrule
    \end{tabular}
\end{table*}

\begin{table}[h]
\vspace{-2mm}
\caption{
    Class-to-image generation performance.  Speedups are calculated on an H100 GPU with a sample batch size of 8. 
    }
\vspace{-1mm}
\centering  \scriptsize
\setlength{\tabcolsep}{4.8pt}
\begin{tabular}{lcccccc}
\toprule
\textbf{Methods} & \textbf{FID $\downarrow$} & \textbf{sFID $\downarrow$} & \textbf{IS $\uparrow$} & \textbf{Precision\%} & \textbf{Recall\%} & \textbf{Speedup} \\
\midrule
\multicolumn{7}{c}{DiT-XL} \\ \cmidrule{1-7}
\rowcolor[gray]{0.9} 
original        & 2.30 & 4.71 & 276.26 & 82.68  & 57.65   & 1.00$\times$ \\
FORA            & 2.45 & 5.44 & 265.94 & 81.21  & 58.36  & 1.57$\times$ \\
Delta-DiT       & 2.47 & 5.61 & 265.33 & 81.05  & \textbf{58.83 } & 1.45$\times$ \\
\cmidrule{1-7} \multicolumn{7}{c}{\ours~ with Cached Inference} \\ \cmidrule{1-7}
\rowcolor[gray]{0.9} 
original & 2.29 & 4.58 & 281.81 & 82.88  & 57.53  & 1.00$\times$ \\
$n=2$ & \textbf{2.31} & \textbf{4.76} & \textbf{277.51} & \textbf{82.52 } & 58.06  & 1.46$\times$ \\
$n=3$  & 2.40 & 4.98 & 272.05 & 82.14  & 57.86  & 1.73$\times$ \\
$n=4$  & 2.54 & 5.31 & 267.34 & 81.60  & 58.31  & \textbf{1.93$\times$} \\
\bottomrule
\end{tabular}
\label{tab:c2i-2}
 \vspace{-2mm}
\end{table}

We extend \ours~to the class-to-image task, where \ours~exceeds DiT-XL with only around 23\% of its training cost (1.6M training steps vs.7M), and we choose the \ours~trained with 2.9M steps for evaluation. Hunyuan-DiT~\cite{li2024hunyuandit} is a powerful text-to-image DiT model featuring LSCs. However, its skip connections have not been fully explored to accelerate generation. We first explore their impact and efficiency potential and integrate it with our caching mechanism without architectural modifications. 
\vspace{-2mm}
\paragraph{Evaluation Details}
To evaluate the generalization of the caching mechanism in \ours~for text-to-image tasks, we use the zero-shot Fréchet Inception Distance (FID) on the MS-COCO~\cite{coco} validation dataset by generating 30k images based on its prompts, following the guidelines of Hunyuan-DiT. Additionally, we also employ Peak Signal-to-Noise Ratio (PSNR), Learned Perceptual Image Patch Similarity (LPIPS), and Structural Similarity Index Measure (SSIM) to evaluate similarity. For the class-to-image task, we evaluate models on ImageNet~\cite{deng2009imagenet} by generating 50k images, and calculating evaluation metrics used in~\citet{dit-xl}.
\vspace{-2mm}
\paragraph{Evaluation Results}
Table \ref{tab:t2i} comprehensively compares Hunyuan-DiT and various caching methods. Notably, our caching mechanism achieves a 2.09$\times$ speedup without any degradation in FID or CLIP scores. Furthermore, it outperforms all other caching methods in terms of PSNR, LPIPS, and SSIM scores, consistently maintaining the highest performance even with a 1.96$\times$ speedup. These findings underscore the robustness and adaptability of our caching mechanism to image generation tasks. Additionally, the unusual improvement in FID and CLIP scores is due to:
\ding{182} In HunyuanDiT's standard evaluation pipeline, FID fluctuates due to zero-shot evaluation and resolution scaling ($1024\rightarrow256$) prior to evaluation.
\ding{183} The CLIP score is fluctuating within a reasonable range (-0.16, 0.2). These fluctuations are also observed in $\Delta$-DiT.
The results on the class-to-image task shown in Table \ref{tab:c2i-2}, \ours~ also achieve the least loss while maintaining higher speedup. 


\subsection{Ablation Studies}
\begin{figure}
    \centering
    \vspace{-2mm}
    \caption{
    The feature dynamics of Latte with LSCs. Differences in features at the same layers across timesteps are evaluated.}
    \vspace{-2mm}
    \includegraphics[width=0.8\linewidth]{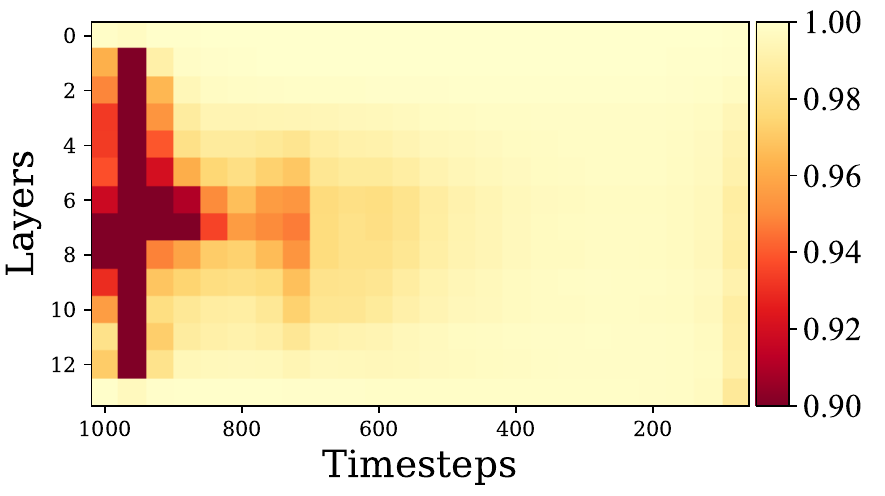}
    \label{fig:heatmap2}
\end{figure}

\paragraph{Identify Dynamic Phase for Better Caching} 
After incorporating LSCs into DiT, we can statically choose the timesteps for caching with feature stability, thus we can identify the dynamic phase by caching similarity shown in Figure~\ref{fig:teaser2}. However, previous work has observed that the denoising process is split into different phases in U-Net diffusion models by~\citet{tgates} and ~\citet{li2023fasterdiffusion}, highlighting the hidden dynamic phases which may lead to better performance.
A heat map visualizing inner feature dynamics within blocks is shown in Figure~\ref{fig:heatmap2}, and significant changes are observed in the early denoising phase.
In Table~\ref{tab:ablation_timestep}, we further validate this observation: under equivalent throughput, caching in the later timesteps (700$\rightarrow$50), outperforms caching in the earlier timesteps (950$\rightarrow$300) significantly. Additionally, we segmented the rapidly changing timesteps and experimented with additional caching ranges. All of these results show that increasing the ratio of the former phase significantly improves caching performance. 
\begin{table}
    \caption{
        Ablation study on different timestep ranges for caching. Total timesteps is 1000. Caching is performed at \textit{caching timesteps}.
    }
    \vspace{-1mm}
    \label{tab:ablation_timestep}
    \setlength{\tabcolsep}{6.7pt}
    \scriptsize \centering
    \begin{tabular}{ccccc}
        \toprule
        \bf Caching Timestep & \textbf{VBench} (\%) $\uparrow$ & \bf PSNR $\uparrow$  & \bf LPIPS $\downarrow$ & \bf SSIM $\uparrow$ \\
        \midrule
        {700$\rightarrow$50}  & \textbf{75.51} & \textbf{29.52} & \textbf{0.06} & \textbf{0.89} \\
        {950$\rightarrow$300} & 75.48 & 20.58          & 0.23          & 0.73         \\
        \midrule
        {800$\rightarrow$50}  & 75.36         & 26.02          & 0.10          & 0.84          \\
        {900$\rightarrow$50}  & 75.24         & 22.13          & 0.19          & 0.76          \\
        \bottomrule
    \end{tabular}
    \vspace{-3mm}
\end{table}

\vspace{-2mm}
\paragraph{Compatibility of \ours~}
As shown in Table~\ref{tab:t2v_a1}, we extend the other text-to-video DiT caching methods to \ours~and observe slight performance improvements. Taking PAB~\cite{zhao2024pab} as an example, it loses 1.15\% less in VBench and achieves noticeably better PSNR and SSIM scores on \ours~, demonstrating the compatibility of \ours~, and that the stable features in \ours~enhance the model robustness and caching efficiency.

\begin{table}
\vspace{-2mm}
    \caption{
        Compatibility of \ours~ with other caching methods. Obvious improvements can be seen especially in PAB.}
    \vspace{-2mm}
    \label{tab:t2v_a1}
    \scriptsize \centering
    \setlength{\tabcolsep}{8.7pt}
    \begin{tabular}{lcccc}
        \toprule
        \bf Method & \textbf{$\Delta$VBench} (\%) $\downarrow$ & \bf PSNR $\uparrow$  & \bf LPIPS $\downarrow$ & \bf SSIM $\uparrow$ \\
        \midrule
        \cmidrule{1-5}
        T-GATE & 0.44 & \textbf{24.09}	& \textbf{0.16}	& 0.78 \\
        $\Delta$-DiT & \textbf{0.12} &	22.64	& 0.17	& \textbf{0.79} \\
        FORA & 0.22 &	23.26 &	0.16 &	0.79 \\
        PAB$_{235}$ & 1.81 &	19.92 &	0.29 &	0.68 \\
        PAB$_{347}$ & 3.52 &	18.98 &	0.34 &	0.65 \\
        PAB$_{469}$ & 3.96 &	18.02 &	0.37 &	0.62 \\
        \bottomrule
    \end{tabular}
    \vspace{-2mm}
\end{table}

\section{Conclusion}\label{sec:conclusion}
\vspace{-1mm}
In this work, we identify the \textit{Dynamic Feature Instability} in vanilla DiT models and introduce \ours~, a Long-Skip-Connections enhanced DiT model to produce more stable features and build training and inference-efficient diffusion transformers. We designed a simple, static, but more efficient caching mechanism for \ours~ to improve inference efficiency in video and image generation tasks. 
By improving the stability of dynamic features, \ours~ unlocks the potential to cache most blocks while maintaining high generation quality. In addition, we guarantee stability and efficiency with theoretical analysis from the perspective of the spectral norm and visualizations.
Furthermore, we identify the dynamic phase of \ours~ for a better caching strategy, demonstrating the effectiveness of caching. We also show that \ours~ is compatible with other caching methods, further extending its applicability.
Overall, we believe that \ours~ provides a simple yet powerful foundation for advancing future research and applications.

\newpage
\section*{Acknowledgement}
During this project, Guanjie Chen is supported by Shanghai
Artificial Intelligence Laboratory.

{
    \small
    \bibliographystyle{ieeenat_fullname}
    \bibliography{main}

@String(CVPR= {IEEE Conf. Comput. Vis. Pattern Recog.})

@String(ICCV= {Int. Conf. Comput. Vis.})

@String(ECCV= {Eur. Conf. Comput. Vis.})

@String(ICLR = {Int. Conf. Learn. Represent.})

@String(CVPR  = {CVPR})

@String(ICCV  = {ICCV})

@String(ECCV  = {ECCV})

@String(ICLR  = {ICLR})

@article{Bao2022AllAW,
  title={All are Worth Words: A ViT Backbone for Diffusion Models},
  author={Fan Bao and Shen Nie and Kaiwen Xue and Yue Cao and Chongxuan Li and Hang Su and Jun Zhu},
  journal={2023 IEEE/CVF Conference on Computer Vision and Pattern Recognition (CVPR)},
  year={2022},
  pages={22669-22679},
}

@article{Chen2023PixArtFT,
  title={PixArt-$\alpha$: Fast Training of Diffusion Transformer for Photorealistic Text-to-Image Synthesis},
  author={Junsong Chen and Jincheng Yu and Chongjian Ge and Lewei Yao and Enze Xie and Yue Wu and Zhongdao Wang and James T. Kwok and Ping Luo and Huchuan Lu and Zhenguo Li},
  journal={ArXiv},
  year={2023},
  volume={abs/2310.00426},}

@article{Podell2023SDXLIL,
  title={SDXL: Improving Latent Diffusion Models for High-Resolution Image Synthesis},
  author={Dustin Podell and Zion English and Kyle Lacey and A. Blattmann and Tim Dockhorn and Jonas Muller and Joe Penna and Robin Rombach},
  journal={ArXiv},
  year={2023},
  volume={abs/2307.01952},
}

@inproceedings{BetkerImprovingIG,
  title={Improving Image Generation with Better Captions},
  author={James Betker and Gabriel Goh and Li Jing and † TimBrooks and Jianfeng Wang and Linjie Li and † LongOuyang and † JuntangZhuang and † JoyceLee and † YufeiGuo and † WesamManassra and † PrafullaDhariwal and † CaseyChu and † YunxinJiao and Aditya Ramesh},
}

@article{su2025thinking,
  title={Thinking with Images for Multimodal Reasoning: Foundations, Methods, and Future Frontiers},
  author={Su, Zhaochen and Xia, Peng and Guo, Hangyu and Liu, Zhenhua and Ma, Yan and Qu, Xiaoye and Liu, Jiaqi and Li, Yanshu and Zeng, Kaide and Yang, Zhengyuan and others},
  journal={arXiv preprint arXiv:2506.23918},
  year={2025}
}

@article{su2025openthinkimg,
  title={OpenThinkIMG: Learning to Think with Images via Visual Tool Reinforcement Learning},
  author={Su, Zhaochen and Li, Linjie and Song, Mingyang and Hao, Yunzhuo and Yang, Zhengyuan and Zhang, Jun and Chen, Guanjie and Gu, Jiawei and Li, Juntao and Qu, Xiaoye and others},
  journal={arXiv preprint arXiv:2505.08617},
  year={2025}
}

@article{Rombach2021HighResolutionIS,
  title={High-Resolution Image Synthesis with Latent Diffusion Models},
  author={Robin Rombach and A. Blattmann and Dominik Lorenz and Patrick Esser and Bj{\"o}rn Ommer},
  journal={2022 IEEE/CVF Conference on Computer Vision and Pattern Recognition (CVPR)},
  year={2021},
  pages={10674-10685},
}

@ARTICLE{2024arXiv241013720P,
       author = {{Polyak}, Adam and {Zohar}, Amit and {Brown}, Andrew and {Tjandra}, Andros and {Sinha}, Animesh and {Lee}, Ann and {Vyas}, Apoorv and {Shi}, Bowen and {Ma}, Chih-Yao and {Chuang}, Ching-Yao and {Yan}, David and {Choudhary}, Dhruv and {Wang}, Dingkang and {Sethi}, Geet and {Pang}, Guan and {Ma}, Haoyu and {Misra}, Ishan and {Hou}, Ji and {Wang}, Jialiang and {Jagadeesh}, Kiran and {Li}, Kunpeng and {Zhang}, Luxin and {Singh}, Mannat and {Williamson}, Mary and {Le}, Matt and {Yu}, Matthew and {Singh}, Mitesh Kumar and {Zhang}, Peizhao and {Vajda}, Peter and {Duval}, Quentin and {Girdhar}, Rohit and {Sumbaly}, Roshan and {Saketh Rambhatla}, Sai and {Tsai}, Sam and {Azadi}, Samaneh and {Datta}, Samyak and {Chen}, Sanyuan and {Bell}, Sean and {Ramaswamy}, Sharadh and {Sheynin}, Shelly and {Bhattacharya}, Siddharth and {Motwani}, Simran and {Xu}, Tao and {Li}, Tianhe and {Hou}, Tingbo and {Hsu}, Wei-Ning and {Yin}, Xi and {Dai}, Xiaoliang and {Taigman}, Yaniv and {Luo}, Yaqiao and {Liu}, Yen-Cheng and {Wu}, Yi-Chiao and {Zhao}, Yue and {Kirstain}, Yuval and {He}, Zecheng and {He}, Zijian and {Pumarola}, Albert and {Thabet}, Ali and {Sanakoyeu}, Artsiom and {Mallya}, Arun and {Guo}, Baishan and {Araya}, Boris and {Kerr}, Breena and {Wood}, Carleigh and {Liu}, Ce and {Peng}, Cen and {Vengertsev}, Dimitry and {Schonfeld}, Edgar and {Blanchard}, Elliot and {Juefei-Xu}, Felix and {Nord}, Fraylie and {Liang}, Jeff and {Hoffman}, John and {Kohler}, Jonas and {Fire}, Kaolin and {Sivakumar}, Karthik and {Chen}, Lawrence and {Yu}, Licheng and {Gao}, Luya and {Georgopoulos}, Markos and {Moritz}, Rashel and {Sampson}, Sara K. and {Li}, Shikai and {Parmeggiani}, Simone and {Fine}, Steve and {Fowler}, Tara and {Petrovic}, Vladan and {Du}, Yuming},
        title = "{Movie Gen: A Cast of Media Foundation Models}",
      journal = {arXiv e-prints},
         year = 2024,
        month = oct,
          eid = {arXiv:2410.13720},
        pages = {arXiv:2410.13720},
          doi = {10.48550/arXiv.2410.13720},
archivePrefix = {arXiv},
       eprint = {2410.13720},
 primaryClass = {cs.CV},
}

@misc{li2024hunyuandit,
      title={Hunyuan-DiT: A Powerful Multi-Resolution Diffusion Transformer with Fine-Grained Chinese Understanding}, 
      author={Zhimin Li and Jianwei Zhang and Qin Lin and Jiangfeng Xiong and Yanxin Long and Xinchi Deng and Yingfang Zhang and Xingchao Liu and Minbin Huang and Zedong Xiao and Dayou Chen and Jiajun He and Jiahao Li and Wenyue Li and Chen Zhang and Rongwei Quan and Jianxiang Lu and Jiabin Huang and Xiaoyan Yuan and Xiaoxiao Zheng and Yixuan Li and Jihong Zhang and Chao Zhang and Meng Chen and Jie Liu and Zheng Fang and Weiyan Wang and Jinbao Xue and Yangyu Tao and Jianchen Zhu and Kai Liu and Sihuan Lin and Yifu Sun and Yun Li and Dongdong Wang and Mingtao Chen and Zhichao Hu and Xiao Xiao and Yan Chen and Yuhong Liu and Wei Liu and Di Wang and Yong Yang and Jie Jiang and Qinglin Lu},
      year={2024},
      eprint={2405.08748},
      archivePrefix={arXiv},
      primaryClass={cs.CV}
}

@article{ma2024latte,
  title={Latte: Latent Diffusion Transformer for Video Generation},
  author={Ma, Xin and Wang, Yaohui and Jia, Gengyun and Chen, Xinyuan and Liu, Ziwei and Li, Yuan-Fang and Chen, Cunjian and Qiao, Yu},
  journal={arXiv preprint arXiv:2401.03048},
  year={2024}
}

@misc{zhao2024pab,
      title={Real-Time Video Generation with Pyramid Attention Broadcast},
      author={Xuanlei Zhao and Xiaolong Jin and Kai Wang and Yang You},
      year={2024},
      eprint={2408.12588},
      archivePrefix={arXiv},
      primaryClass={cs.CV},
      url={https://arxiv.org/abs/2408.12588},
}

@software{opensora,
  author = {Zangwei Zheng and Xiangyu Peng and Tianji Yang and Chenhui Shen and Shenggui Li and Hongxin Liu and Yukun Zhou and Tianyi Li and Yang You},
  title = {Open-Sora: Democratizing Efficient Video Production for All},
  month = {March},
  year = {2024},
  url = {https://github.com/hpcaitech/Open-Sora}
}

@misc{openai2024sora,
  title={Sora: Creating video from text},
  author={OpenAI},
  year={2024},
  howpublished={\url{https://openai.com/sora}},
}

@software{pku_opensora_plan,
  author       = {PKU-Yuan Lab and Tuzhan AI etc.},
  title        = {Open-Sora-Plan},
  month        = apr,
  year         = 2024,
  publisher    = {GitHub},
  doi          = {10.5281/zenodo.10948109},
  url          = {https://doi.org/10.5281/zenodo.10948109}
}

@article{Peebles2022ScalableDM,
  title={Scalable Diffusion Models with Transformers},
  author={William S. Peebles and Saining Xie},
  journal={2023 IEEE/CVF International Conference on Computer Vision (ICCV)},
  year={2022},
  pages={4172-4182},
  url={https://api.semanticscholar.org/CorpusID:254854389}
}

@article{DeepCache,
  title={DeepCache: Accelerating Diffusion Models for Free},
  author={Xinyin Ma and Gongfan Fang and Xinchao Wang},
  journal={2024 IEEE/CVF Conference on Computer Vision and Pattern Recognition (CVPR)},
  year={2023},
  pages={15762-15772},
  url={https://api.semanticscholar.org/CorpusID:265609065}
}

@article{li2023fasterdiffusion,
  title={Faster diffusion: Rethinking the role of unet encoder in diffusion models},
  author={Li, Senmao and Hu, Taihang and Khan, Fahad Shahbaz and Li, Linxuan and Yang, Shiqi and Wang, Yaxing and Cheng, Ming-Ming and Yang, Jian},
  journal={arXiv preprint arXiv:2312.09608},
  year={2023}
}

@inproceedings{wimbauer2024blockcache,
  title={Cache me if you can: Accelerating diffusion models through block caching},
  author={Wimbauer, Felix and Wu, Bichen and Schoenfeld, Edgar and Dai, Xiaoliang and Hou, Ji and He, Zijian and Sanakoyeu, Artsiom and Zhang, Peizhao and Tsai, Sam and Kohler, Jonas and others},
  booktitle={Proceedings of the IEEE/CVF Conference on Computer Vision and Pattern Recognition},
  pages={6211--6220},
  year={2024}
}

@article{so2023frdiff,
  title={FRDiff: Feature Reuse for Exquisite Zero-shot Acceleration of Diffusion Models},
  author={So, Junhyuk and Lee, Jungwon and Park, Eunhyeok},
  journal={arXiv preprint arXiv:2312.03517},
  year={2023}
}

@article{zhang2024cross,
  title={Cross-attention makes inference cumbersome in text-to-image diffusion models},
  author={Zhang, Wentian and Liu, Haozhe and Xie, Jinheng and Faccio, Francesco and Shou, Mike Zheng and Schmidhuber, J{\"u}rgen},
  journal={arXiv preprint arXiv:2404.02747},
  year={2024}
}

@article{yang2024cogvideox,
  title={CogVideoX: Text-to-Video Diffusion Models with An Expert Transformer},
  author={Yang, Zhuoyi and Teng, Jiayan and Zheng, Wendi and Ding, Ming and Huang, Shiyu and Xu, Jiazheng and Yang, Yuanming and Hong, Wenyi and Zhang, Xiaohan and Feng, Guanyu and others},
  journal={arXiv preprint arXiv:2408.06072},
  year={2024}
}

@article{ffs,
  author       = {Andreas R{\"{o}}ssler and
                  Davide Cozzolino and
                  Luisa Verdoliva and
                  Christian Riess and
                  Justus Thies and
                  Matthias Nie{\ss}ner},
  title        = {FaceForensics: {A} Large-scale Video Dataset for Forgery Detection
                  in Human Faces},
  journal      = {CoRR},
  volume       = {abs/1803.09179},
  year         = {2018},
  url          = {http://arxiv.org/abs/1803.09179},
  eprinttype    = {arXiv},
  eprint       = {1803.09179},
  bibsource    = {dblp computer science bibliography, https://dblp.org}
}

@inproceedings{sky,
  author       = {Wei Xiong and
                  Wenhan Luo and
                  Lin Ma and
                  Wei Liu and
                  Jiebo Luo},
  title        = {Learning to Generate Time-Lapse Videos Using Multi-Stage Dynamic Generative
                  Adversarial Networks},
  booktitle    = {2018 {IEEE} Conference on Computer Vision and Pattern Recognition,
                  {CVPR} 2018, Salt Lake City, UT, USA, June 18-22, 2018},
  pages        = {2364--2373},
  publisher    = {Computer Vision Foundation / {IEEE} Computer Society},
  year         = {2018},
  url          = {http://openaccess.thecvf.com/content\_cvpr\_2018/html/Xiong\_Learning\_to\_Generate\_CVPR\_2018\_paper.html},
  doi          = {10.1109/CVPR.2018.00251},
  bibsource    = {dblp computer science bibliography, https://dblp.org}
}

@article{ucf101,
  author       = {Khurram Soomro and
                  Amir Roshan Zamir and
                  Mubarak Shah},
  title        = {{UCF101:} {A} Dataset of 101 Human Actions Classes From Videos in
                  The Wild},
  journal      = {CoRR},
  volume       = {abs/1212.0402},
  year         = {2012},
  url          = {http://arxiv.org/abs/1212.0402},
  eprinttype    = {arXiv},
  eprint       = {1212.0402},
  bibsource    = {dblp computer science bibliography, https://dblp.org}
}

@inproceedings{taichi,
  author       = {Aliaksandr Siarohin and
                  St{\'{e}}phane Lathuili{\`{e}}re and
                  Sergey Tulyakov and
                  Elisa Ricci and
                  Nicu Sebe},
  editor       = {Hanna M. Wallach and
                  Hugo Larochelle and
                  Alina Beygelzimer and
                  Florence d'Alch{\'{e}}{-}Buc and
                  Emily B. Fox and
                  Roman Garnett},
  title        = {First Order Motion Model for Image Animation},
  booktitle    = {Advances in Neural Information Processing Systems 32: Annual Conference
                  on Neural Information Processing Systems 2019, NeurIPS 2019, December
                  8-14, 2019, Vancouver, BC, Canada},
  pages        = {7135--7145},
  year         = {2019},
  url          = {https://proceedings.neurips.cc/paper/2019/hash/31c0b36aef265d9221af80872ceb62f9-Abstract.html},
  bibsource    = {dblp computer science bibliography, https://dblp.org}
}

@inproceedings{webvid,
  title={Frozen in time: A joint video and image encoder for end-to-end retrieval},
  author={Bain, Max and Nagrani, Arsha and Varol, G{\"u}l and Zisserman, Andrew},
  booktitle={Proceedings of the IEEE/CVF international conference on computer vision},
  pages={1728--1738},
  year={2021}
}

@article{vimeo,
  author       = {Yaohui Wang and
                  Xinyuan Chen and
                  Xin Ma and
                  Shangchen Zhou and
                  Ziqi Huang and
                  Yi Wang and
                  Ceyuan Yang and
                  Yinan He and
                  Jiashuo Yu and
                  Peiqing Yang and
                  Yuwei Guo and
                  Tianxing Wu and
                  Chenyang Si and
                  Yuming Jiang and
                  Cunjian Chen and
                  Chen Change Loy and
                  Bo Dai and
                  Dahua Lin and
                  Yu Qiao and
                  Ziwei Liu},
  title        = {{LAVIE:} High-Quality Video Generation with Cascaded Latent Diffusion
                  Models},
  journal      = {CoRR},
  volume       = {abs/2309.15103},
  year         = {2023},
  url          = {https://doi.org/10.48550/arXiv.2309.15103},
  doi          = {10.48550/ARXIV.2309.15103},
  eprinttype    = {arXiv},
  eprint       = {2309.15103},
  bibsource    = {dblp computer science bibliography, https://dblp.org}
}

@InProceedings{VBench,
    author    = {Huang, Ziqi and He, Yinan and Yu, Jiashuo and Zhang, Fan and Si, Chenyang and Jiang, Yuming and Zhang, Yuanhan and Wu, Tianxing and Jin, Qingyang and Chanpaisit, Nattapol and Wang, Yaohui and Chen, Xinyuan and Wang, Limin and Lin, Dahua and Qiao, Yu and Liu, Ziwei},
    title     = {VBench: Comprehensive Benchmark Suite for Video Generative Models},
    booktitle = {Proceedings of the IEEE/CVF Conference on Computer Vision and Pattern Recognition (CVPR)},
    month     = {June},
    year      = {2024},
    pages     = {21807-21818}
}

@inproceedings{lpips,
  title={The unreasonable effectiveness of deep features as a perceptual metric},
  author={Zhang, Richard and Isola, Phillip and Efros, Alexei A and Shechtman, Eli and Wang, Oliver},
  booktitle={Proceedings of the IEEE conference on computer vision and pattern recognition},
  pages={586--595},
  year={2018}
}

@article{ssim,
  author       = {Zhou Wang and
                  Alan C. Bovik},
  title        = {A universal image quality index},
  journal      = {{IEEE} Signal Process. Lett.},
  volume       = {9},
  number       = {3},
  pages        = {81--84},
  year         = {2002},
  url          = {https://doi.org/10.1109/97.995823},
  doi          = {10.1109/97.995823},
  bibsource    = {dblp computer science bibliography, https://dblp.org}
}

@article{fid,
  author       = {Gaurav Parmar and
                  Richard Zhang and
                  Jun{-}Yan Zhu},
  title        = {On Buggy Resizing Libraries and Surprising Subtleties in {FID} Calculation},
  journal      = {CoRR},
  volume       = {abs/2104.11222},
  year         = {2021},
  url          = {https://arxiv.org/abs/2104.11222},
  eprinttype    = {arXiv},
  eprint       = {2104.11222},
  bibsource    = {dblp computer science bibliography, https://dblp.org}
}

@article{fvd,
  author       = {Thomas Unterthiner and
                  Sjoerd van Steenkiste and
                  Karol Kurach and
                  Rapha{\"{e}}l Marinier and
                  Marcin Michalski and
                  Sylvain Gelly},
  title        = {Towards Accurate Generative Models of Video: {A} New Metric {\&}
                  Challenges},
  journal      = {CoRR},
  volume       = {abs/1812.01717},
  year         = {2018},
  url          = {http://arxiv.org/abs/1812.01717},
  eprinttype    = {arXiv},
  eprint       = {1812.01717},
  bibsource    = {dblp computer science bibliography, https://dblp.org}
}

@article{tgates,
  author       = {Wentian Zhang and
                  Haozhe Liu and
                  Jinheng Xie and
                  Francesco Faccio and
                  Mike Zheng Shou and
                  J{\"{u}}rgen Schmidhuber},
  title        = {Cross-Attention Makes Inference Cumbersome in Text-to-Image Diffusion
                  Models},
  journal      = {CoRR},
  volume       = {abs/2404.02747},
  year         = {2024},
  url          = {https://doi.org/10.48550/arXiv.2404.02747},
  doi          = {10.48550/ARXIV.2404.02747},
  eprinttype    = {arXiv},
  eprint       = {2404.02747},
  bibsource    = {dblp computer science bibliography, https://dblp.org}
}

@article{delta,
  title={Delta-DiT: A Training-Free Acceleration Method Tailored for Diffusion Transformers},
  author={Chen, Pengtao and Shen, Mingzhu and Ye, Peng and Cao, Jianjian and Tu, Chongjun and Bouganis, Christos-Savvas and Zhao, Yiren and Chen, Tao},
  journal={arXiv preprint arXiv:2406.01125},
  year={2024}
}

@article{fora,
  title={Fora: Fast-forward caching in diffusion transformer acceleration},
  author={Selvaraju, Pratheba and Ding, Tianyu and Chen, Tianyi and Zharkov, Ilya and Liang, Luming},
  journal={arXiv preprint arXiv:2407.01425},
  year={2024}
}

@inproceedings{ddim,
  author       = {Jiaming Song and
                  Chenlin Meng and
                  Stefano Ermon},
  title        = {Denoising Diffusion Implicit Models},
  booktitle    = {9th International Conference on Learning Representations, {ICLR} 2021,
                  Virtual Event, Austria, May 3-7, 2021},
  publisher    = {OpenReview.net},
  year         = {2021},
  url          = {https://openreview.net/forum?id=St1giarCHLP},
  bibsource    = {dblp computer science bibliography, https://dblp.org}
}

@article{Ronneberger2015UNetCN,
  title={U-Net: Convolutional Networks for Biomedical Image Segmentation},
  author={Olaf Ronneberger and Philipp Fischer and Thomas Brox},
  journal={ArXiv},
  year={2015},
  volume={abs/1505.04597},
}

@article{sauer2023adversarial,
  title={Adversarial diffusion distillation},
  author={Sauer, Axel and Lorenz, Dominik and Blattmann, Andreas and Rombach, Robin},
  journal={arXiv preprint arXiv:2311.17042},
  year={2023}
}

@article{chen2024qdit,
  title={Q-dit: Accurate post-training quantization for diffusion transformers},
  author={Chen, Lei and Meng, Yuan and Tang, Chen and Ma, Xinzhu and Jiang, Jingyan and Wang, Xin and Wang, Zhi and Zhu, Wenwu},
  journal={arXiv preprint arXiv:2406.17343},
  year={2024}
}

@inproceedings{yin2024one,
  title={One-step diffusion with distribution matching distillation},
  author={Yin, Tianwei and Gharbi, Micha{\"e}l and Zhang, Richard and Shechtman, Eli and Durand, Fredo and Freeman, William T and Park, Taesung},
  booktitle={Proceedings of the IEEE/CVF Conference on Computer Vision and Pattern Recognition},
  pages={6613--6623},
  year={2024}
}

@inproceedings{ddpm,
  author       = {Jonathan Ho and
                  Ajay Jain and
                  Pieter Abbeel},
  editor       = {Hugo Larochelle and
                  Marc'Aurelio Ranzato and
                  Raia Hadsell and
                  Maria{-}Florina Balcan and
                  Hsuan{-}Tien Lin},
  title        = {Denoising Diffusion Probabilistic Models},
  booktitle    = {Advances in Neural Information Processing Systems 33: Annual Conference
                  on Neural Information Processing Systems 2020, NeurIPS 2020, December
                  6-12, 2020, virtual},
  year         = {2020},
  url          = {https://proceedings.neurips.cc/paper/2020/hash/4c5bcfec8584af0d967f1ab10179ca4b-Abstract.html},
  bibsource    = {dblp computer science bibliography, https://dblp.org}
}

@inproceedings{stylegenv,
    title={Generating Videos with Dynamics-aware Implicit Generative Adversarial Networks},
    author={Sihyun Yu and Jihoon Tack and Sangwoo Mo and Hyunsu Kim and Junho Kim and Jung-Woo Ha and Jinwoo Shin},
    booktitle={International Conference on Learning Representations},
    year={2022},
    url={https://openreview.net/forum?id=Czsdv-S4-w9}
}

@inproceedings{coco,
  author       = {Tsung{-}Yi Lin and
                  Michael Maire and
                  Serge J. Belongie and
                  James Hays and
                  Pietro Perona and
                  Deva Ramanan and
                  Piotr Doll{\'{a}}r and
                  C. Lawrence Zitnick},
  editor       = {David J. Fleet and
                  Tom{\'{a}}s Pajdla and
                  Bernt Schiele and
                  Tinne Tuytelaars},
  title        = {Microsoft {COCO:} Common Objects in Context},
  booktitle    = {Computer Vision - {ECCV} 2014 - 13th European Conference, Zurich,
                  Switzerland, September 6-12, 2014, Proceedings, Part {V}},
  series       = {Lecture Notes in Computer Science},
  volume       = {8693},
  pages        = {740--755},
  publisher    = {Springer},
  year         = {2014},
  url          = {https://doi.org/10.1007/978-3-319-10602-1\_48},
  doi          = {10.1007/978-3-319-10602-1\_48},
  bibsource    = {dblp computer science bibliography, https://dblp.org}
}

@inproceedings{dit-xl,
  author       = {William Peebles and
                  Saining Xie},
  title        = {Scalable Diffusion Models with Transformers},
  booktitle    = {{IEEE/CVF} International Conference on Computer Vision, {ICCV} 2023,
                  Paris, France, October 1-6, 2023},
  pages        = {4172--4182},
  publisher    = {{IEEE}},
  year         = {2023},
  url          = {https://doi.org/10.1109/ICCV51070.2023.00387},
  doi          = {10.1109/ICCV51070.2023.00387},
  bibsource    = {dblp computer science bibliography, https://dblp.org}
}

@inproceedings{deng2009imagenet,
  title={Imagenet: A large-scale hierarchical image database},
  author={Deng, Jia and Dong, Wei and Socher, Richard and Li, Li-Jia and Li, Kai and Fei-Fei, Li},
  booktitle={2009 IEEE conference on computer vision and pattern recognition},
  pages={248--255},
  year={2009},
  organization={Ieee}
}

@inproceedings{zhang2024pia,
  title={Pia: Your personalized image animator via plug-and-play modules in text-to-image models},
  author={Zhang, Yiming and Xing, Zhening and Zeng, Yanhong and Fang, Youqing and Chen, Kai},
  booktitle={Proceedings of the IEEE/CVF Conference on Computer Vision and Pattern Recognition},
  pages={7747--7756},
  year={2024}
}

@article{SpectralNormRegularization,
  author       = {Yuichi Yoshida and
                  Takeru Miyato},
  title        = {Spectral Norm Regularization for Improving the Generalizability of
                  Deep Learning},
  journal      = {CoRR},
  volume       = {abs/1705.10941},
  year         = {2017},
  url          = {http://arxiv.org/abs/1705.10941},
  eprinttype    = {arXiv},
  eprint       = {1705.10941},
  bibsource    = {dblp computer science bibliography, https://dblp.org}
}

@inproceedings{SpectralNormalizationGAN,
  author       = {Takeru Miyato and
                  Toshiki Kataoka and
                  Masanori Koyama and
                  Yuichi Yoshida},
  title        = {Spectral Normalization for Generative Adversarial Networks},
  booktitle    = {6th International Conference on Learning Representations, {ICLR} 2018,
                  Vancouver, BC, Canada, April 30 - May 3, 2018, Conference Track Proceedings},
  publisher    = {OpenReview.net},
  year         = {2018},
  url          = {https://openreview.net/forum?id=B1QRgziT-},
  bibsource    = {dblp computer science bibliography, https://dblp.org}
}

@article{adacache,
  title={Adaptive caching for faster video generation with diffusion transformers},
  author={Kahatapitiya, Kumara and Liu, Haozhe and He, Sen and Liu, Ding and Jia, Menglin and Zhang, Chenyang and Ryoo, Michael S and Xie, Tian},
  journal={arXiv preprint arXiv:2411.02397},
  year={2024}
}

@article{teacache,
  title={Timestep Embedding Tells: It's Time to Cache for Video Diffusion Model},
  author={Liu, Feng and Zhang, Shiwei and Wang, Xiaofeng and Wei, Yujie and Qiu, Haonan and Zhao, Yuzhong and Zhang, Yingya and Ye, Qixiang and Wan, Fang},
  journal={arXiv preprint arXiv:2411.19108},
  year={2024}
}

@article{scalelsc,
  title={Scalelong: Towards more stable training of diffusion model via scaling network long skip connection},
  author={Huang, Zhongzhan and Zhou, Pan and Yan, Shuicheng and Lin, Liang},
  journal={Advances in Neural Information Processing Systems},
  volume={36},
  pages={70376--70401},
  year={2023}
}

@article{zhao2024dynamic,
  title={Dynamic diffusion transformer},
  author={Zhao, Wangbo and Han, Yizeng and Tang, Jiasheng and Wang, Kai and Song, Yibing and Huang, Gao and Wang, Fan and You, Yang},
  journal={arXiv preprint arXiv:2410.03456},
  year={2024}
}

@article{yao2025fasterdit,
  title={Fasterdit: Towards faster diffusion transformers training without architecture modification},
  author={Yao, Jingfeng and Wang, Cheng and Liu, Wenyu and Wang, Xinggang},
  journal={Advances in Neural Information Processing Systems},
  volume={37},
  pages={56166--56189},
  year={2025}
}

@inproceedings{u_dit,
  author       = {Yuchuan Tian and
                  Zhijun Tu and
                  Hanting Chen and
                  Jie Hu and
                  Chao Xu and
                  Yunhe Wang},
  editor       = {Amir Globersons and
                  Lester Mackey and
                  Danielle Belgrave and
                  Angela Fan and
                  Ulrich Paquet and
                  Jakub M. Tomczak and
                  Cheng Zhang},
  title        = {U-DiTs: Downsample Tokens in U-Shaped Diffusion Transformers},
  booktitle    = {Advances in Neural Information Processing Systems 38: Annual Conference
                  on Neural Information Processing Systems 2024, NeurIPS 2024, Vancouver,
                  BC, Canada, December 10 - 15, 2024},
  year         = {2024},
  url          = {http://papers.nips.cc/paper\_files/paper/2024/hash/5d2e24df9cfaad3189833b819c40b392-Abstract-Conference.html},
  bibsource    = {dblp computer science bibliography, https://dblp.org}
}
}
\clearpage
\setcounter{page}{1}
\maketitlesupplementary
\section{Limitations}
Despite achieving significant speedups, \ours~ inherits limitations from its DiT foundation. including high computational demands, reliance on large-scale training data, and quadratic complexity that hinders high-resolution generation. Furthermore, \ours~ introduces a marginal parameter overhead (5.5\% for class-to-image, 3.5\% for text-to-video) and extra GPU memory (scales with the batch size $\mathcal{N}$), requiring an additional $\mathcal{N}\times0.37\%$ for the class-to-image model and $\mathcal{N}\times4.55\%$ for the text-to-video model. These models remain feasible for deployment with proper $\mathcal{N}$.
\section{Supplementary Experiments}
\paragraph{Analysis of Block Selection for Caching}
To identify the optimal block for caching, we analyze the feature similarity across timesteps in the final three blocks of Latte-T2V (Table \ref{tab:connectionselection}). Our analysis reveals that caching at block 27, the last DiT block, yields the best performance. This block, which is connected to block 0 via the primary Long Skip Connection (LSC), not only exhibits maximum feature similarity but also enables the highest speedup. These results demonstrate the superior caching efficiency achieved by \ours.
\begin{table}[H]
    \scriptsize \centering
    \caption{
    Connection selection for caching in the text-to-video task. The best metrics are emphasized in \textbf{bold}.
}
    \label{tab:connectionselection}
    \setlength{\tabcolsep}{3pt}
    \begin{tabular}{ccccccc}
        \toprule
        LSC & Cached & Similarity(\%)$\uparrow$ & VBench(\%) $\uparrow$ & PSNR $\uparrow$  & LPIPS $\downarrow$ & SSIM $\uparrow$\\
        \midrule
         0 & [1,26] & \textbf{99.88} & \textbf{75.51} &	29.52 & \textbf{0.06} & \textbf{0.89}\\
        1 & [2,25] & 99.86 & 75.44 &	\textbf{29.65} & \textbf{0.06} & \textbf{0.89}\\
        2 & [3,24] & 99.66 & 75.49 &	29.44 & \textbf{0.06} & 0.88\\
        \bottomrule
    \end{tabular}
\end{table}

\paragraph{Comparison in Few-Step Generation}
We benchmarked \ours~against a vanilla DiT in a few-step text-to-video generation scenario, using a 5 and 10-step DDIM scheduler instead of the standard 50 steps. As shown in Table \ref{tab:t2vFewSteps}, while the vanilla DiT holds a slight advantage in the SSIM metric, \ours~achieves superior visual quality and semantic consistency. These results again validate the balanced and robust performance of our architecture, even in a highly accelerated, few-step setting.
\begin{table}
    \scriptsize \centering
    \caption{
    Text-to-video generation performance under the few-step scenario. The best metrics are emphasized in \textbf{bold}.
}
    \label{tab:t2vFewSteps}
    \setlength{\tabcolsep}{4pt}
    \begin{tabular}{clcccc}
        \toprule
        NFE & Model & $\Delta$VBench(\%) & PSNR $\uparrow$  & LPIPS $\downarrow$ & SSIM $\uparrow$\\
        \midrule
         \multirow{2}{*}{\textbf{10 steps}} & \textbf{Latte} & $\downarrow$1.73 &	15.96 &	0.47 & \textbf{0.61}\\
        & \textbf{\ours} & \textbf{$\downarrow$0.94} &	\textbf{16.18} &	\textbf{0.43} & 0.59\\
        \cmidrule{1-6}
        \multirow{2}{*}{\textbf{5 steps}} & \textbf{Latte} & $\downarrow$9.40 &	12.94 &	0.73 & \textbf{0.53}\\
        & \textbf{\ours} & \textbf{$\downarrow$8.55} &	\textbf{13.17} &	\textbf{0.69} & 0.50\\
        \bottomrule
    \end{tabular}
\end{table}

\section{Detailed Proof of Theorems}
Consider an ideal denoising diffusion transformer $M$ with $L$ identical blocks, where the denoising capability induces the following fundamental properties:
\begin{itemize}[leftmargin=*]
\setlength{\itemsep}{0pt} 
\item \textit{Noise Reduction Invariance}: Each transformer block $\mathcal{T}_l$ strictly reduces noise magnitude of input $h$. This directly implies the Jacobian spectral norm ($\sigma_{\max}$) constraint:
\vspace{-1mm}
    \begin{equation}
    \sigma_{\max}\left( \frac{\partial \mathcal{T}_l}{\partial h} \right) \triangleq \gamma_{l} < 1
    \end{equation}
    
    \item \textit{Transformer Blocks Homogeneity}: Identical noise reduction ratio across layers
    $\gamma_l = \gamma,\quad \forall l \in \{1,...,L\}$. Thus, the complete model satisfies
\vspace{-1mm}
\begin{equation}
    \sigma_{\max}(M) = \prod_{l=1}^L \gamma = \gamma^L \ll 1
\end{equation}
\end{itemize}
\subsection{Theorem1}
\textit{The spectral norm of the Jacobian matrix of DiT with Long-Skip-Connections is controlled tighter than that of Vanilla DiT $M$, making the Skip-DiT model more robust, numerically stable, and capable of converging faster.}
\begin{proof}
Define layer transformations for \( L/2 < l \leq L \) and their Jacobian matrices, 
where \( \mathcal{T} \) denotes the Transformer block and \( 0 < \alpha < 1 \):
\begin{align*}
h_{l+1}^{\text{vanilla}} = \mathcal{T}(h_l^{\text{vanilla}}), h_{l+1}^{\text{skip}} = (1-\alpha) \cdot \mathcal{T}(h_l^{\text{skip}}) + \alpha \cdot h_{L-l}^{\text{skip}}.
\end{align*}
The corresponding Jacobian matrices $J$ are:
\begin{align*}
J_{l}^{\text{vanilla}} = \frac{\partial \mathcal{T}\left(h_l\right)}{\partial h_l}, J_{l}^{\text{skip}} = (1-\alpha) \cdot \frac{\partial \mathcal{T}\left(h_l\right)}{\partial h_l} + \alpha \cdot \frac{\partial h_{L-l}}{\partial h_l}.
\end{align*}
Applying subadditivity and submultiplicativity of spectral norms $\sigma_{max}$:
\begin{align}
\sigma_{\max}(J_{l}^{\mathrm{skip}}) &\leq (1-\alpha)\gamma + \alpha\gamma^{2l-L}.
\end{align}
Given $\gamma < 1$ and $2l-L \geq 1$ for $l > L/2$, we establish the layer-wise bound:
\begin{equation}
\sigma_{\max}(J_{l}^{\mathrm{skip}}) < \gamma = \sigma_{\max}(J_{l}^{\mathrm{vanilla}}).
\end{equation}
For the full model Jacobian $J = \prod_{l=1}^{L} J_l$, the spectral norms satisfy:
\begin{align}
\sigma_{\max}(J^{\mathrm{skip}}) &\leq \prod_{l=0}^{L-1} \left[(1-\alpha)\gamma + \alpha\gamma^{2l-L}\right] \nonumber \\
                                  &< \prod_{l=0}^{L-1} \gamma = \gamma^L = \sigma_{\max}(J^{\mathrm{vanilla}}).
\end{align}
This spectral radius reduction attenuates perturbation growth through $\| \delta h_{L} \| \leq (\sigma_{\max}(J))^L \| \delta h_0 \|$, stabilizes gradient flow via $\| \nabla_{\theta}\mathcal{L} \| \leq \sigma_{\max}(J) \| \nabla_{h_L}\mathcal{L} \|$, and improves the Lipschitz constant $\mathrm{Lip}(M_{\mathrm{skip}}) < \mathrm{Lip}(M_{\mathrm{vanilla}})$.
\end{proof}

\subsection{Theorem2}
\textit{Under feature reuse with interval $\tau$, Skip-DiT achieves tighter perturbation bounds than Vanilla DiT, enabling larger allowable reuse intervals while maintaining error tolerance $\epsilon_{\mathrm{max}}$}
\begin{proof}
Let $h_{t}$ be the reused feature with error $\epsilon_t$, $\delta$ the perturbation bound per step, $x$ the input, and $\theta$ the model parameters. Through the differential mean-value theorem and error propagation:
\begin{align}
\|f(x;\theta+\Delta\theta) - f(x;\theta)\| &\leq \sup_{\theta'}\left\|\frac{\partial f}{\partial \theta}\right\|\|\Delta\theta\| \triangleq \delta \label{eq:delta_def} \\
\epsilon_{t+1}=\| h_{t+1}-h_t \| &\leq \mathrm{Lip}(M)\epsilon_t + \delta \label{eq:error_prop}
\end{align} 
From Theorem~\ref{theorem1}, the Lipschitz constants satisfy:
\begin{equation}
\mathrm{Lip}(M_{\mathrm{skip}}) = \sigma_{\max}(J^{\mathrm{skip}}) < \gamma^L = \mathrm{Lip}(M_{\mathrm{vanilla}}) \label{eq:lip_relation}
\end{equation}
The perturbation bounds inherit:
\begin{equation}
\delta_{\mathrm{skip}} = C_{\mathrm{skip}}\|\Delta\theta\| < C_{\mathrm{vanilla}}\|\Delta\theta\| = \delta_{\mathrm{vanilla}} \label{eq:delta_relation} 
\end{equation}
where $C_{\cdot}$ bounds the parameter-to-output Jacobian. 
\begin{align*}
\left\|\frac{\partial f_{\mathrm{skip}}}{\partial \theta}\right\| 
&= \left\|\prod_{l=1}^L \frac{\partial h_l}{\partial \theta}\right\|
\leq \prod_{l=1}^L \left\|\frac{\partial h_l}{\partial \theta}\right\| \\
&= \prod_{l=1}^L \left[(1-\alpha)\gamma + \alpha\gamma^{2l-L}\right] \cdot C_{\mathrm{base}} \\
&< \prod_{l=1}^L \gamma \cdot C_{\mathrm{base}} = C_{\mathrm{vanilla}}
\end{align*}

For $T$-step feature reuse, cumulative error develops as:
\begin{equation}
\epsilon_T \leq \frac{\mathrm{Lip}(M)^T - 1}{\mathrm{Lip}(M) - 1}\delta \label{eq:cumulative_error}
\end{equation}
Given $\sigma_{\max}(J^{\mathrm{skip}}) < \gamma^L$ and $\delta_{\mathrm{skip}} < \delta_{\mathrm{vanilla}}$, the maximal interval $\tau$ satisfies:
\begin{equation}
\frac{\sigma_{\max}(J^{\mathrm{skip}})^\tau - 1}{\sigma_{\max}(J^{\mathrm{skip}}) - 1}\delta_{\mathrm{skip}} = \frac{\gamma^{L\tau} - 1}{\gamma^L - 1}\delta_{\mathrm{vanilla}} = \epsilon_{\mathrm{max}} \label{eq:interval_sup}
\end{equation}
Solving \eqref{eq:interval_sup} yields $\tau_{\mathrm{skip}} > \tau_{\mathrm{vanilla}}$ under identical $\epsilon_{\mathrm{max}}$, proving Skip-DiT permits larger reuse intervals.
\end{proof}

\section{Class-to-image Generation Experiments}
\label{apx:class2image}
\citet{dit-xl} proposed the first diffusion model based on the transformer architecture, and it outperforms all prior diffusion models on the class conditional ImageNet~\cite{deng2009imagenet} 512$\times$512 and 256$\times$256 benchmarks. We add skip connections to its largest model, DiT-XL, to get \ours. We train \ours~on class conditional ImageNet with resolution 256$\times$256 from scratch with completely the same experimental settings as DiT-XL, and far exceeds DiT-XL with only around 38\% of its training cost.

\vspace{-1mm}
\paragraph{Training of \ours}
We add long-skip-connections in DiT-XL and train \ours~ for 2.9M steps on 8 A100 GPUs, compared to 7M steps for DiT-XL, which also uses 8 A100 GPUs. The datasets and other training settings remain identical to those used for DiT-XL, and we utilize the official training code of DiT-XL\footnote{https://github.com/facebookresearch/DiT}. The performance comparison is presented in Table \ref{tab:c2v_training}, which demonstrates that \ours~ significantly outperforms DiT-XL while requiring only 23\% of its training steps (1.6M vs. 7M), highlighting the training efficiency and effectiveness of \ours.


\begin{table}[t]
\caption{
    Comparison of training efficiency between DiT-XL and \ours~. Images are generated with a 250-step DDPM solver. The term \textit{cfg} refers to classifier-free guidance scales, where metrics for \textit{cfg=1.0} are computed without classifier-free guidance. The best metrics are highlighted in \textbf{bold}. \ours~ significantly exceeds DiT-XL/2 with much less training steps.
}
\centering
\scriptsize
\begin{tabular}{l c c c c c c}
\toprule
\textbf{Model} & \textbf{Steps} & \textbf{FID $\downarrow$} & \textbf{sFID $\downarrow$} & \textbf{IS $\uparrow$} & \textbf{Precision $\uparrow$} & \textbf{Recall $\uparrow$} \\
\midrule
DiT-XL/2 & 7000k & 9.49 & 7.17 & 122.49 & 0.67 & 0.68 \\
\midrule
\multirow{6}{*}{Skip-DiT} & 400k & 13.46 & 5.83 & 87.45 & 0.67 & 0.63 \\
 & 800k & 10.13 & 5.87 & 108.28 & 0.68 & 0.65 \\
 & 1600k & 9.07 & 6.26 & 119.38 & 0.68 & 0.67 \\
 & 2200k & 8.59 & 6.41 & 124.74 & 0.68 & 0.67 \\
 & 2500k & 8.41 & 6.30 & 125.16 & 0.68 & 0.67 \\
 & 2900k & 8.37 & 6.50 & 127.63 & 0.68 & 0.68 \\
\bottomrule
\end{tabular}
\label{tab:c2v_training}
\end{table}

\vspace{-1mm}
\paragraph{Accelerating Evaluation }

\begin{table}[h]

\caption{
    Class-to-image generation performance. The definition of the cache step $n$ follows that in Table 4. Images are generated with a 250-step DDPM solver. Speedups are calculated on an H100 GPU with a sample batch size of 8. $n=i$ indicates caching the high-level features in $x_t$ for reuse during the inference of $x_{t-1}, x_{t-2}, \dots, x_{t-n+1}$. The term \textit{cfg} refers to classifier-free guidance scales, where metrics for \textit{cfg=1.0} are computed without classifier-free guidance. We highlight baseline DiT models (without caching) in \colorbox{gray!20}{grey} and the best metrics in \textbf{bold}.
}
\centering  \scriptsize
\setlength{\tabcolsep}{4.8pt}
\begin{tabular}{lcccccc}
\toprule
\textbf{Methods} & \textbf{FID $\downarrow$} & \textbf{sFID $\downarrow$} & \textbf{IS $\uparrow$} & \textbf{Precision\%} & \textbf{Recall\%} & \textbf{Speedup} \\
\midrule
\multicolumn{7}{c}{\textbf{\textit{cfg=1.5}}} \\ \cmidrule{1-7}
\rowcolor[gray]{0.9} 
DiT-XL        & 2.30 & 4.71 & 276.26 & 82.68  & 57.65   & 1.00$\times$ \\
FORA            & 2.45 & 5.44 & 265.94 & 81.21  & 58.36  & 1.57$\times$ \\
Delta-DiT       & 2.47 & 5.61 & 265.33 & 81.05  & 58.83  & 1.45$\times$ \\
Faster-Diff & 4.96 & 10.19 & 223.21 & 75.21  & \textbf{59.28} & 1.42$\times$ \\
\cmidrule{1-7} \multicolumn{7}{c}{\ours~with Cached Inference} \\ \cmidrule{1-7}
\rowcolor[gray]{0.9} 
\ours        & 2.29 & 4.58 & 281.81 & 82.88  & 57.53  & 1.00$\times$ \\
$n=2$ & \textbf{2.31} & \textbf{4.76} & \textbf{277.51} & \textbf{82.52 } & 58.06  & 1.46$\times$ \\
$n=3$  & 2.40 & 4.98 & 272.05 & 82.14  & 57.86  & 1.73$\times$ \\
$n=4$  & 2.54 & 5.31 & 267.34 & 81.60  & 58.31  & \textbf{1.93$\times$} \\
\midrule

\multicolumn{7}{c}{ \textbf{\textit{cfg=1.0}}}  \\ \cmidrule{1-7}
\rowcolor[gray]{0.9}
DiT-XL        & 9.49 & 7.17 & 122.49 & 66.66  & 67.69   & 1.00$\times$ \\
FORA            & 11.72 & 9.27 & 113.01 & 64.46  & 67.69  & 1.53$\times$ \\
Delta-DiT       & 12.03 & 9.68 & 111.86 & 64.57  & 67.53  & 1.42$\times$ \\
Faster-Diff & 22.98 & 18.09 & 80.41 & 56.53  & 67.41 & 1.42$\times$ \\
\cmidrule{1-7} \multicolumn{7}{c}{\ours~with Cached Inference} \\ \cmidrule{1-7}
\rowcolor[gray]{0.9} 
\ours & 8.37 & 6.50 & 127.63 & 68.06  & 67.89  & 1.00$\times$ \\
$n=2$ & \textbf{9.25} & \textbf{7.09} & \textbf{123.57} & \textbf{67.32 } & 67.40  & 1.46$\times$ \\
$n=3$  & 10.18 & 7.72 & 119.60 & 66.53  & \textbf{67.84 } & 1.71$\times$ \\
$n=4$  & 11.37 & 8.49 & 116.01 & 65.73  & 67.32  & \textbf{1.92$\times$} \\
\bottomrule
\end{tabular}
\label{tab:c2i}
\end{table}

We evaluate \ours~ and compare its performance against two other caching methods: $\Delta$-DiT and FORA. As shown in Table \ref{tab:c2i}, \ours~achieves a 1.46$\times$ speedup with only a minimal FID loss of $0.02$ when the classifier-free guidance scale is set to 1.5, compared to the 7–8$\times$ larger losses observed with $\Delta$-DiT and FORA. Moreover, even with a 1.9$\times$ acceleration, \ours~ performs better than the other caching methods. These findings further confirm the effectiveness of \ours~ for class-to-image tasks.




\section{Evaluation Details}
\vspace{-1mm} \paragraph{VBench~\cite{VBench}} is a novel evaluation framework for video generation models. It breaks down video generation assessment to 16 dimensions from video quality and condition consistency: subject consistency, background consistency, temporal flickering, motion smoothness, dynamic degree, aesthetic quality, imaging quality, object class, multiple objects, human action, color, spatial relationship, scene, temporal style, appearance style, overall consistency.

\vspace{-1mm} \paragraph{Peak Signal-to-Noise Ratio (PSNR)} measures generated visual content quality by comparing a processed version $\mathbf{v}$ to the original reference $\mathbf{v}_{r}$ by:
\begin{align}
    \mathrm{PSNR} = 10 \times \log_{10} (\frac{R^2}{\texttt{MSE}(\mathbf{v},\mathbf{v}_r)})
\end{align}
where $R$ is the maximum possible pixel value, and $\texttt{MSE}(\cdot,\cdot)$ calculates the Mean Squared Error between original and processed images or videos. Higher PSNR indicates better reconstruction quality. However, PSNR does not always correlate with human perception and is sensitive to pixel-level changes.

\vspace{-1mm} \paragraph{Structural Similarity Index Measure (SSIM)} is a perceptual metric that evaluates image quality by considering luminance, contrast, and structure:
\begin{align}
    \mathrm{SSIM} = [l(\mathbf{v},\mathbf{v}_r)]^\alpha \cdot [c(\mathbf{v},\mathbf{v}_r)]^\beta \cdot [s(\mathbf{v},\mathbf{v}_r\mathbf{v}_r)]^\gamma
\end{align}
where $\alpha,\beta,\gamma$ are weights for luminance, contrast, and structure quality, where luminance comparison is $l(x,y) = \frac{2\mu_{\mathbf{v}}\mu_{\mathbf{v}_r} + C_1}{\mu_{\mathbf{v}}^2 + \mu_{\mathbf{v}_r}^2 + C_1} $, contrast comparison is
$c(x,y) = \frac{2\sigma_{\mathbf{v}}\sigma_{\mathbf{v}_r} + C_2}{\sigma_{\mathbf{v}}^2 + \sigma_{\mathbf{v}_r}^2 + C_2}$, and structure comparison is $s(x,y) = \frac{\sigma_{xy} + C_3}{\sigma_{\mathbf{v}}\sigma_{\mathbf{v}_r} + C_3}$, with $C$ denoting numerical stability coefficients. SSIM scores range from -1 to 1, where 1 means identical visual content.

\vspace{-1mm} \paragraph{Learned Perceptual Image Patch Similarity (LPIPS)} is  a deep learning-based metric that measures perceptual similarity using L2-Norm of visual features $v \in \mathbb{R}^{H\times W\times C}$ extracted from pretrained CNN $\mathcal{F}(\cdot)$. LPIPS captures semantic similarities and is therefore more robust to small geometric transformations than PSNR and SSIM.
\begin{align}
    \mathrm{LPIPS} = \frac{1}{H W} \sum_{h,w} ||\mathcal{F}(v_r) - \mathcal{F}(v)||_2^2
\end{align}
\vspace{-1mm} \paragraph{Fr\'{e}chet Inception Distance (FID) and Fr\'{e}chet Video Distance (FVD)} FID measures the quality and diversity of generated images by computing distance between feature distributions of reference $\mathcal{N}(\mu_r,\Sigma_r)$ and generated images $\mathcal{N}(\mu,\Sigma)$ using inception architecture CNNs, where $\mu,\Sigma$ are mean and covariance of features.
\begin{align}
\mathrm{FID} = ||\mu_r - \mu||^2 + Tr(\Sigma_r + \Sigma - 2(\Sigma_r\Sigma)^{1/2})
\end{align}
 FVD is a video extension of FID. Lower FID and FVD indicate higher generation quality.

\begin{table}[th]
    \caption{
    Comparison of our caching method with the faster sampler DDIM. \ourscache~ is evaluated with a 250-steps DDPM and compared to the DDIM sampler under similar throughput. Baseline DiT models (without caching) are highlighted in \colorbox{gray!20}{grey}, and the best metrics are indicated in \textbf{bold}. Notably, DDIM outperforms the 250-step DDPM in the UCF101 task for both Latte and \ours~. \ourscache~ denotes \ours~ with cached inference.
    }
    \label{tab:c2v_a2}
    \scriptsize \centering
    \setlength{\tabcolsep}{2pt}
    \resizebox{\linewidth}{!}{
    \begin{tabular}{clcccccccccccc}
    \toprule
    \multirow{2}{*}{Method} & \multicolumn{2}{c}{\texttt{UCF101}} & \multicolumn{2}{c}{\texttt{FFS}} & \multicolumn{2}{c}{\texttt{Sky}} & \multicolumn{2}{c}{\texttt{Taichi}} \\
    & FVD $\downarrow$ & FID $\downarrow$ & FVD $\downarrow$ & FID $\downarrow$ & FVD $\downarrow$ & FID $\downarrow$ & FVD $\downarrow$ & FID $\downarrow$ \\
    \midrule
    \rowcolor[gray]{0.9}Latte & 165.04 &	23.75 &	28.88 &	5.36 &	49.46 &	11.51 &	166.84 & 11.57 \\
    \rowcolor[gray]{0.9}\ours & 173.70 &	22.95 &	20.62 &	4.32 &	49.22 &	12.05 &	163.03 & 13.55 \\
        \cmidrule{1-9}
        
        \ours$_{n=2}$ & 165.60 & \textbf{22.73} & \textbf{23.55} &	\textbf{4.49} &	\textbf{51.13} &	\textbf{12.66} &	\textbf{167.54} &	\textbf{13.89}  \\
        DDIM+${\texttt{\ours}}$ & \textbf{134.22} &	24.60	& 37.28 &	6.48	& 86.39	& 13.67	& 343.97	& 21.01 \\
        DDIM+${\texttt{Latte}}$ &146.78	& 23.06	& 39.10	& 6.47	& 78.38	& 13.73	& 321.97 &	21.86 \\
        \cmidrule{1-9}

        \ours$_{n=3}$ & 169.37	& \textbf{22.47} &	\textbf{26.76} &	\textbf{4.75} &	\textbf{54.17} &	\textbf{13.11} &	\textbf{179.43} & \textbf{14.53}  \\
        DDIM+${\texttt{\ours}}$ & \textbf{139.52}	 & 24.71	& 39.20 &	6.49	& 90.62	& 13.80	& 328.47	& 21.33 \\
        DDIM$+{\texttt{Latte}}$ &148.46	& 23.41	& 41.00	& 6.54	& 74.39	& 14.20	& 327.22 &	22.96 \\
        
        \bottomrule
    \end{tabular}}
\end{table}

\section{Scheduler Selection for text-to-video Tasks}
For the text-to-video generation task, all the videos generated for evaluation are sampled with 50 steps DDIM~\cite{DDIM}, which is the default setting used in Latte.
In the class-to-video generation tasks, vanilla Latte uses 250-step DDPM \cite{ddpm} as the default solver for class-to-video tasks, which we adopt for all tasks except UCF101. For UCF101, we employ 50-step DDIM \cite{DDIM}, as it outperforms 250-step DDPM on both Latte and \ours. Table~\ref{tab:c2v_a2} highlights this phenomenon, showing our methods consistently outperform DDPM-250 under comparable throughput, except for UCF101, where DDIM performs better than 250 steps DDPM.  In the text-to-image task, we choose 50-step DDIM for sampling, and for the class-to-image task, we choose 250-steps DDPM.
 
\section{Implementation of Other Caching Methods}
\vspace{-1mm} \paragraph{DeepCache}
DeepCache \cite{DeepCache} is a training-free caching method designed for U-Net-based diffusion models, leveraging the inherent temporal redundancy in sequential denoising steps. It utilizes the skip connections of the U-Net to reuse high-level features while updating low-level features efficiently. \ours~ shares significant similarities with DeepCache but extends the method to DiT models. Specifically, we upgrade traditional DiT models to \ours~ and cache them using \ours~. In the work of DeepCache, two key caching decisions are introduced:
(1) \texttt{N}: the number of steps for reusing cached high-level features. Cached features are computed once and reused for the next \texttt{N-1} steps.
(2) The layer at which caching is performed. For instance, caching at the first layer ensures that only the first and last layers of the U-Net are recomputed.
In \ours, we adopt these two caching strategies and additionally account for the timesteps to cache, addressing the greater complexity of DiT models compared to U-Net-based diffusion models. For all tasks except the class-to-image task, caching is performed at the first layer, whereas for the class-to-image task, it is applied at the third layer.

\vspace{-1mm} \paragraph{$\Delta$-DiT}
$\Delta$-DiT \cite{delta} is a training-free caching method designed for image-generating DiT models. Instead of caching the feature maps directly, it uses the offsets of features as cache objects to preserve input information. This approach is based on the observation that the front blocks of DiT are responsible for generating the image outlines, while the rear blocks focus on finer details. A hyperparameter \( b \) is introduced to denote the boundary between the outline and detail generation stages. When \( t \leq b \), $\Delta$-Cache is applied to the rear blocks; when \( t > b \), it is applied to the front blocks. The number of cached blocks is represented by \( N_{c} \).
While this caching method was initially designed for image generation tasks, we extend it to video generation tasks. In video generation, we observe significant degradation in performance when caching the rear blocks, so we restrict caching to the front blocks during the outline generation stage. For Hunyuan-DiT \cite{li2024hunyuandit}, we cache the middle blocks due to the U-shaped transformer architecture. Detailed configurations are provided in Table~\ref{tab:delta-config}.

\begin{table}[htbp]
\centering \scriptsize
\caption{Configuration details for $\Delta$-Cache in different models and tasks. \textit{t2v} denotes text-to-video, \textit{c2v} denotes class-to-video, \textit{t2i} denotes text-to-image, and \textit{c2i} denotes class-to-image.}
\begin{tabular}{lccccc}
\toprule
\textbf{$\Delta$-DiT} & \textbf{Task} & \textbf{Diffusion steps} & \textbf{$b$} & \textbf{All layers} & \textbf{$N_c$} \\
\midrule
Latte & \textit{t2v}    & 50  & 12 & 28 & 21 \\
Latte & \textit{c2v}    & 250 & 60 & 14 & 10 \\
Hunyuan & \textit{t2i} & 50  & 12 & 28 & 18 \\
DiT-XL{/}2 & \textit{c2i} & 250 & 60 & 28 & 21 \\
\bottomrule
\end{tabular}
\label{tab:delta-config}
\end{table}

\vspace{-1mm} \paragraph{PAB}
PAB (Pyramid Attention Broadcast) \cite{zhao2024pab} is one of the most promising caching methods designed for real-time video generation. The method leverages the observation that attention differences during the diffusion process follow a U-shaped pattern, broadcasting attention outputs to subsequent steps in a pyramid-like manner. Different broadcast ranges are set for three types of attention—spatial, temporal, and cross-attention—based on their respective differences. $PAB_{\alpha\beta\gamma}$ denotes the broadcast ranges for spatial ($\alpha$), temporal ($\beta$), and cross ($\gamma$) attentions.

In this work, we use the official implementation of PAB for text-to-video tasks on Latte and adapt the caching method to other tasks in-house. For the class-to-video task, where cross-attention is absent, $PAB_{\alpha\beta}$ refers to the broadcast ranges of spatial ($\alpha$) and temporal ($\beta$) attentions. In the text-to-image task, which lacks temporal attention, $PAB_{\alpha\beta}$ instead denotes the broadcast ranges of spatial ($\alpha$) and cross ($\beta$) attentions. We do not apply PAB to the class-to-image task, as it involves only spatial attention.

\begin{table}[h]
\centering
\scriptsize
\caption{Configuration details for T-GATE in different settings. \textit{t2v} denotes text-to-video, \textit{t2i} denotes text-to-image.}
\begin{tabular}{lcccc}
\toprule
\textbf{T-GATE} & \textbf{Task} & \textbf{Diffusion steps} & \textbf{m} & \textbf{k} \\
\midrule
Latte & \textit{t2v} & 50 & 20 & 2 \\
Hunyuan-DiT  & \textit{t2i} & 50  & 20 & 2 \\
\bottomrule
\end{tabular}
\label{tab:t-gate}
\end{table}
\vspace{-5mm}

\begin{figure*}[h]
    \centering
    \includegraphics[width=\linewidth]{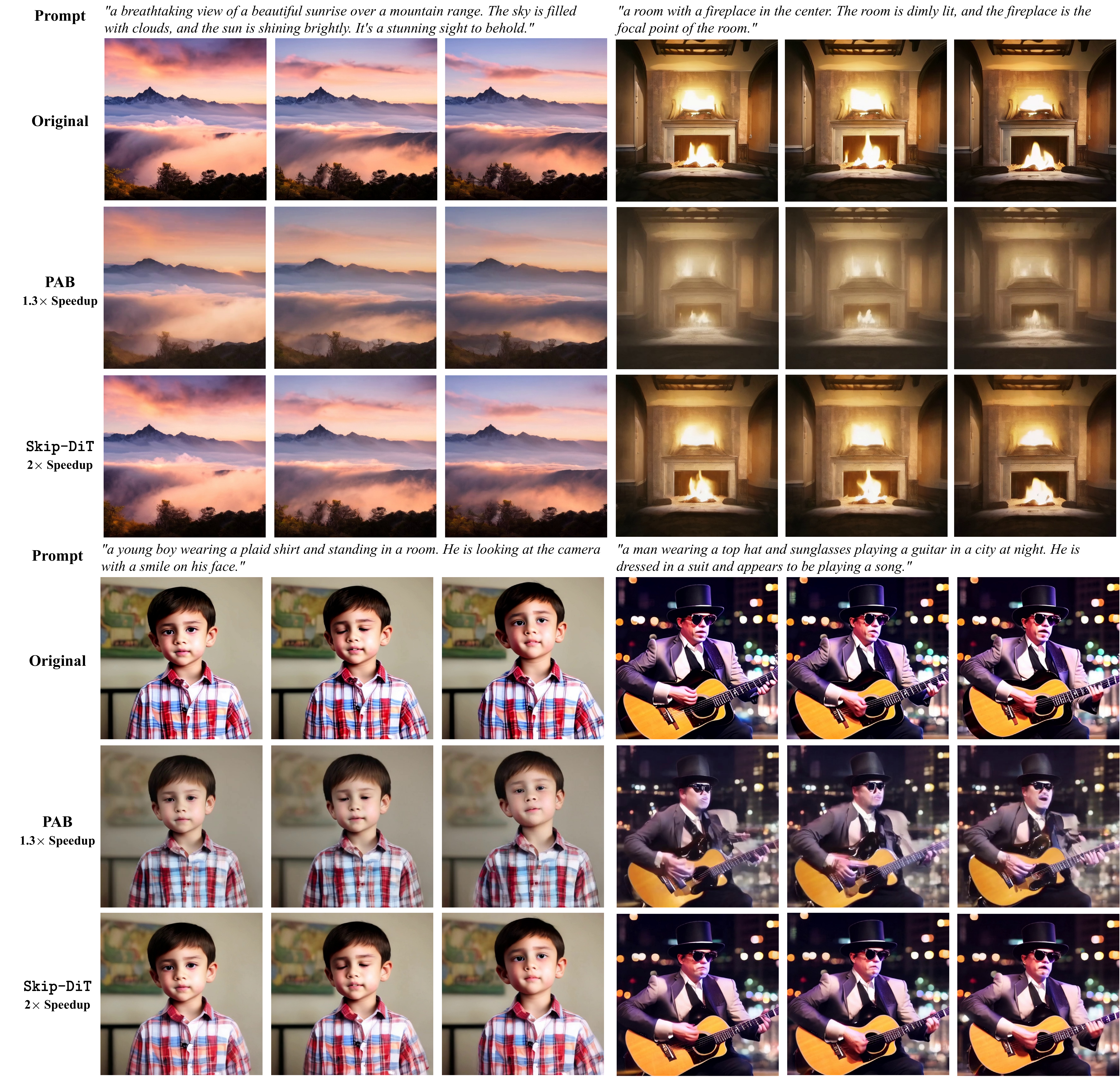}
    \caption{Qualitative results of text-to-video generation. We present \ours~, PAB$_{469}$, and the original model. The frames are randomly sampled from the generated video.}

    \label{fig:case_t2v}
\end{figure*}

\begin{figure*}[h]
    \centering
    \includegraphics[width=\linewidth]{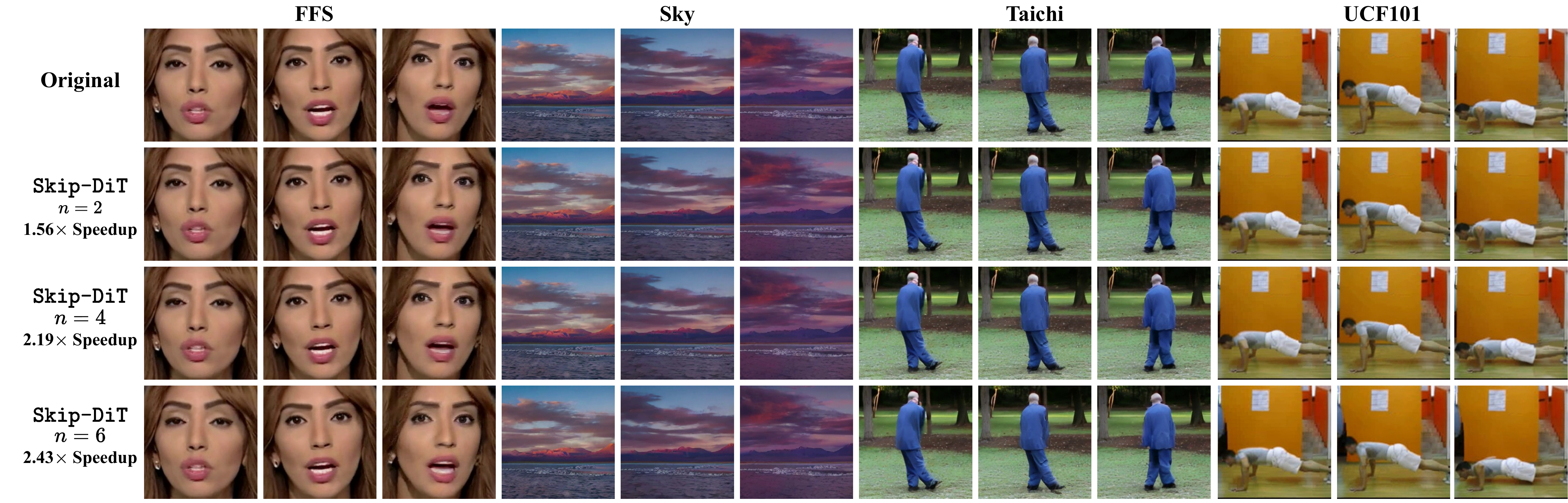}
    \caption{Qualitative results of class-to-video generation. We present the original video generation model and \ours~with different caching steps $n$. The frames are randomly sampled from the generated video.}
    \label{fig:case_c2v}
\end{figure*}

\begin{figure*}[h]
    \centering
    \includegraphics[width=0.9\linewidth]{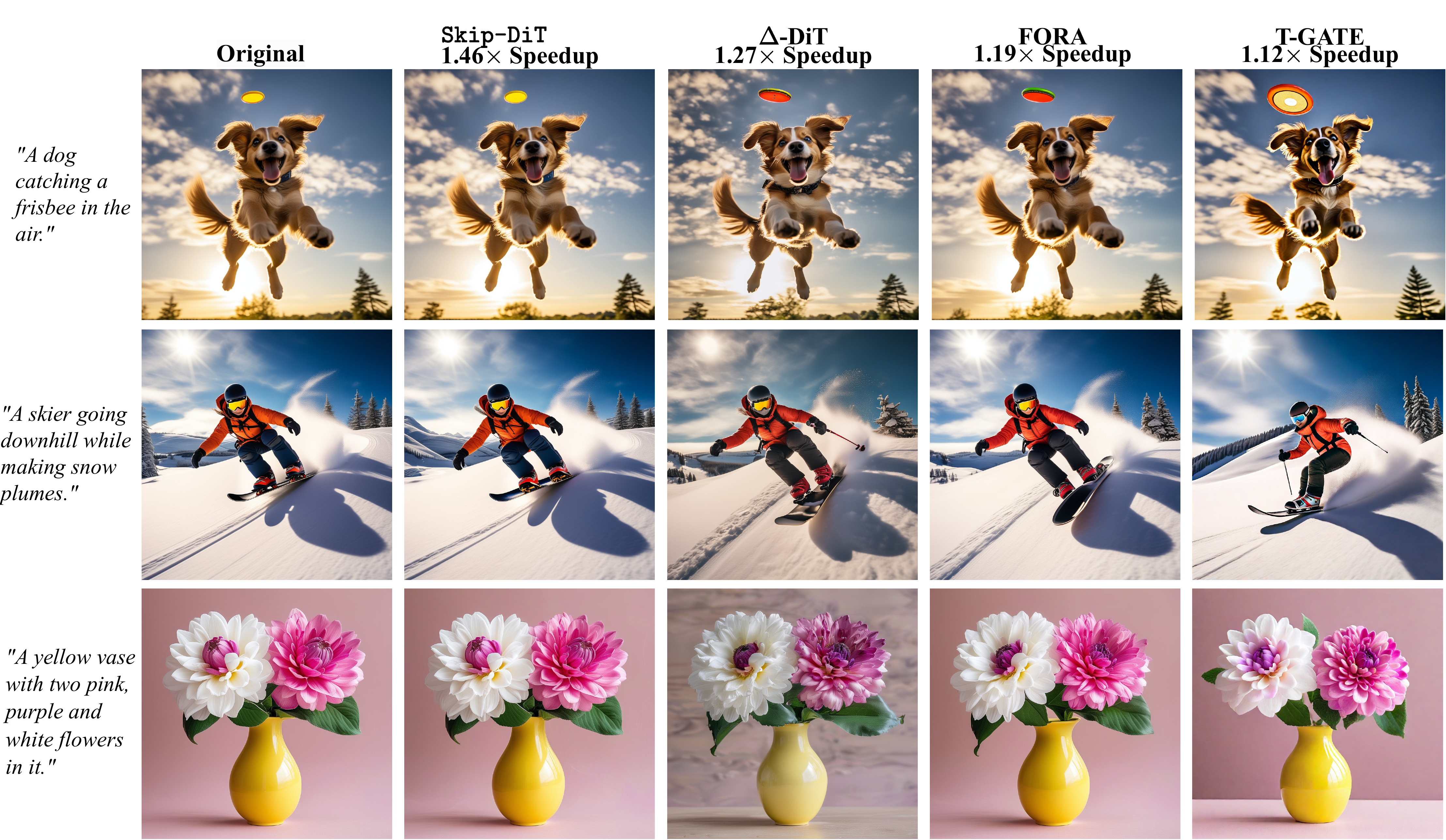}
    \caption{Qualitative results of text-to-image generation. We present \ours~, $\Delta$-DiT, FORA, T-GATE, and the original model. }
    \label{fig:case_t2i}
\end{figure*}

\begin{figure*}[h]
    \centering
    \includegraphics[width=\linewidth]{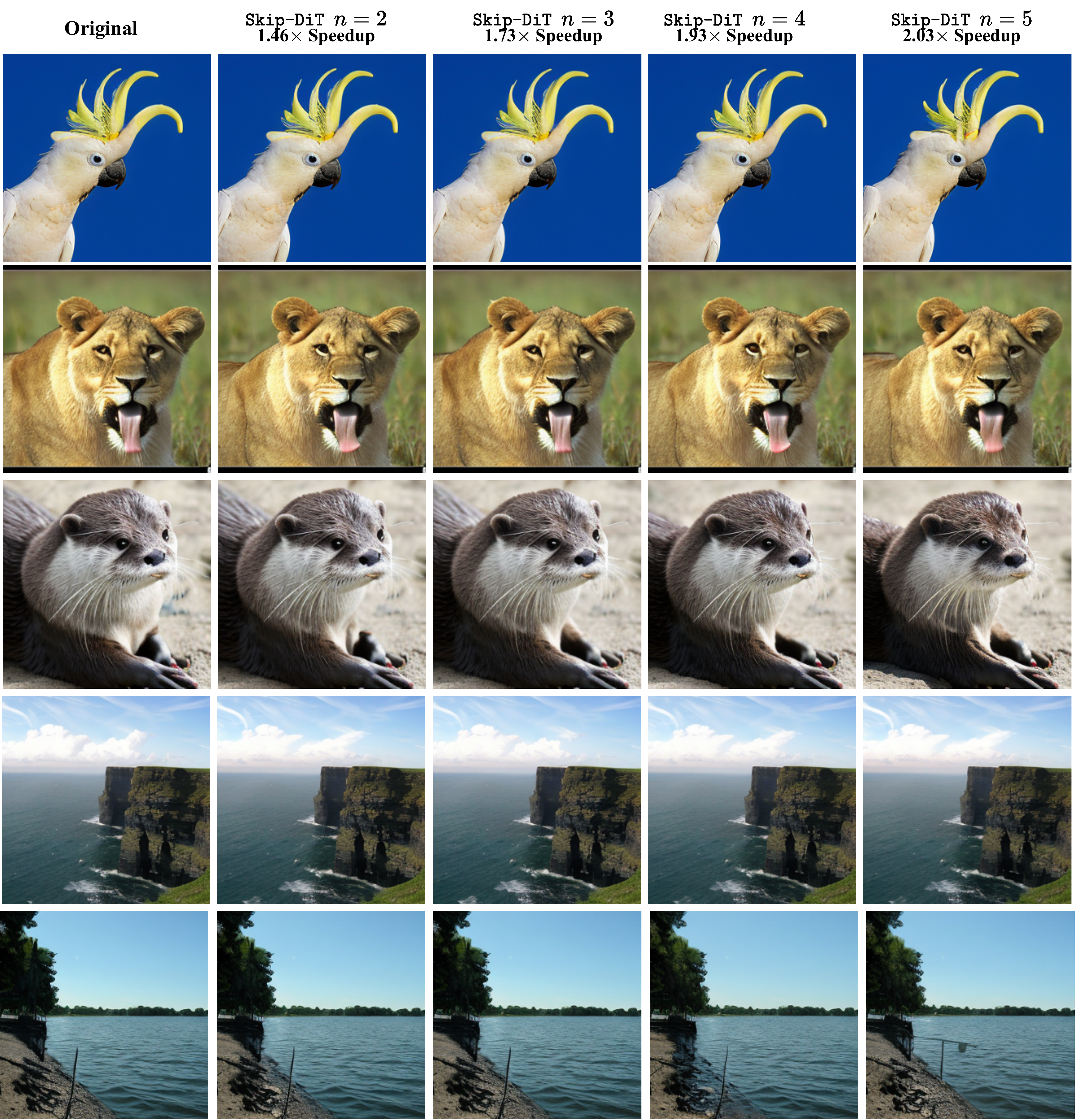}
    \caption{Qualitative results of class-to-image generation. We present the original image generation model and \ours~with different caching steps $n$.}
    \label{fig:case_c2i}
\end{figure*}

\vspace{-1mm} \paragraph{T-Gates}
T-Gates divide the diffusion process into two phases: (1) the Semantics-Planning Phase and (2) the Fidelity-Improving Phase. In the first phase, self-attention is computed and reused every $k$ step. In the second phase, cross-attention is cached using a caching mechanism. The hyperparameter $m$  determines the boundary between these two phases. For our implementation, we use the same hyperparameters as PAB \cite{zhao2024pab}. Detailed configurations are provided in Table \ref{tab:t-gate}.

\vspace{-1mm} \paragraph{FORA}
FORA (Fast-Forward Caching) \cite{fora} stores and reuses intermediate outputs from attention and MLP layers across denoising steps. However, in the original FORA paper, features are cached in advance before the diffusion process. We do not adopt this approach, as it is a highly time-consuming process. Instead, in this work, we skip the “Initialization” step in FORA and calculate the features dynamically during the diffusion process.

\vspace{-1mm} \paragraph{AdaCache}
AdaCache~\cite{adacache} identifies the feature sample-variance in the DiT, and proposes to predict the feature similarity of the current timestep to decide whether to cache. The codebook for Latte, which records the threshold is not released. We provide a version of codebook which almost reproduce the performance in the paper. The codebook is as follows:
$\{0.08: 3, 0.10: 2, 0.12: 1\}$

\vspace{-1mm} \paragraph{TeaCache}
Same as AdaCache, TeaCache\cite{teacache} also discovers the feature instability in DiT. Different from AdaCache, TeaCache finds the relationship between input and output similarity and employ high-order polynomial functions to predict the similarity of the current timestep. We use the official codebase of it and choose the slow-caching strategy to evaluate its upper-bound.
\section{Case Study}
\vspace{-1mm} \paragraph{Video Generation} In Figure~\ref{fig:case_t2v}, we showcase the generated video frames from text prompts with \ours~, PAB, and comparing them to the original model. From generating portraits to scenery, \ours~ with caching consistently demonstrates better visual fidelity along with faster generation speeds. Figure~\ref{fig:case_c2v} presents class-to-video generation examples with \ours~with varying caching steps $\in \{2,4,6\}$. By comparing the output of \ours~ with cache to standard output, we see \ours~maintains good generation quality across different caching steps. 
 
\vspace{-1mm} \paragraph{Image Generation} Figure~\ref{fig:case_t2i} compares qualitative results of \ours~compared to other caching-based acceleration methods ($\Delta$-DiT, FORA, T-GATE) on Hunyuan-DiT. In Figure~\ref{fig:case_c2i}, \ours~show distinct edges in higher speedup and similarity to the original generation, while other baselines exist with different degrees of change in details such as color, texture, and posture. Similarly, we present \ours~with varying caching steps in Figure~\ref{fig:case_c2i}, showing that with more steps cached, it still maintains high fidelity to the original generation.

\end{document}